\documentclass[lettersize,journal]{IEEEtran}
\usepackage{amsmath,amsfonts}
\usepackage{algorithmic}
\usepackage{algorithm}
\usepackage{array}
\usepackage{subcaption}
\usepackage{textcomp}
\usepackage{stfloats}
\usepackage{url}
\usepackage{verbatim}
\usepackage{graphicx}
\usepackage{booktabs}
\usepackage{multirow}
\usepackage{booktabs}
\usepackage{cite}
\usepackage{pifont}
\usepackage{tabularx}

\begin{document}

\title{Toward Equitable Low-Carbon Mobility: Fairness-Aware Demand Prediction for Expanding Bike-Sharing Systems}

\author{Yixuan Zhao,~
        Man Luo$^{*}$%
\thanks{$^{*}$Corresponding author.}
\thanks{Y. Zhao and M. Luo are with the Department of Computer Science, University of Exeter, Exeter EX4 4QF, U.K. (e-mail: \{yz776, m.luo\}@exeter.ac.uk).}
}

\markboth{IEEE Transactions on Knowledge and Data Engineering}%
{Zhao \MakeLowercase{\textit{et al.}}: Fairness-Aware Demand Prediction for Expanding Bike-Sharing Systems}

\maketitle

\begin{abstract}

Bike-sharing systems constitute an important component of low-carbon urban mobility, but their continued expansion presents two closely related challenges. Newly deployed stations lack historical ridership records, creating a discrepancy between training and inference for graph-based models applied to evolving network topologies in which nodes have heterogeneous feature availability. A further challenge arises from the structural biases embedded in historical demand observations. Systematically lower ridership in low-income neighborhoods may reflect inadequate access to cycling infrastructure rather than inherently weak latent demand. Models trained directly on such observations can therefore reproduce existing spatial inequalities, leading deployment strategies to direct green transportation resources toward already advantaged communities and further exacerbate mobility inequity. To address these challenges, we propose \textbf{FairGIN}, a fairness-aware graph neural network for demand prediction in expanding bike-sharing systems. FairGIN comprises three coordinated components. \textit{Expansion-Simulated Increment Training} introduces a graph augmentation mechanism that stochastically simulates cold-start expansion scenarios during training, thereby reducing the distribution discrepancy encountered when new stations are added. \textit{Attention-Based Knowledge Transfer} combines station-adaptive temperature scaling with orthogonal embedding alignment, enabling data-sparse new stations to selectively acquire representations from data-rich existing stations. \textit{Fairness-Aware Optimization} incorporates income-stratified regularization and an equity-calibrated deployment scoring function to translate equitable demand prediction into more inclusive station placement decisions. Experiments on the NYC and Seattle mobility systems demonstrate that FairGIN achieves state-of-the-art predictive accuracy across diverse network expansion scenarios while substantially reducing income-based disparities without compromising overall system efficiency.

\end{abstract}


\section{Introduction}

\begin{figure}[t]
    \centering
    \begin{minipage}[b]{.24\textwidth}
        \centering
        \includegraphics[width=\linewidth]{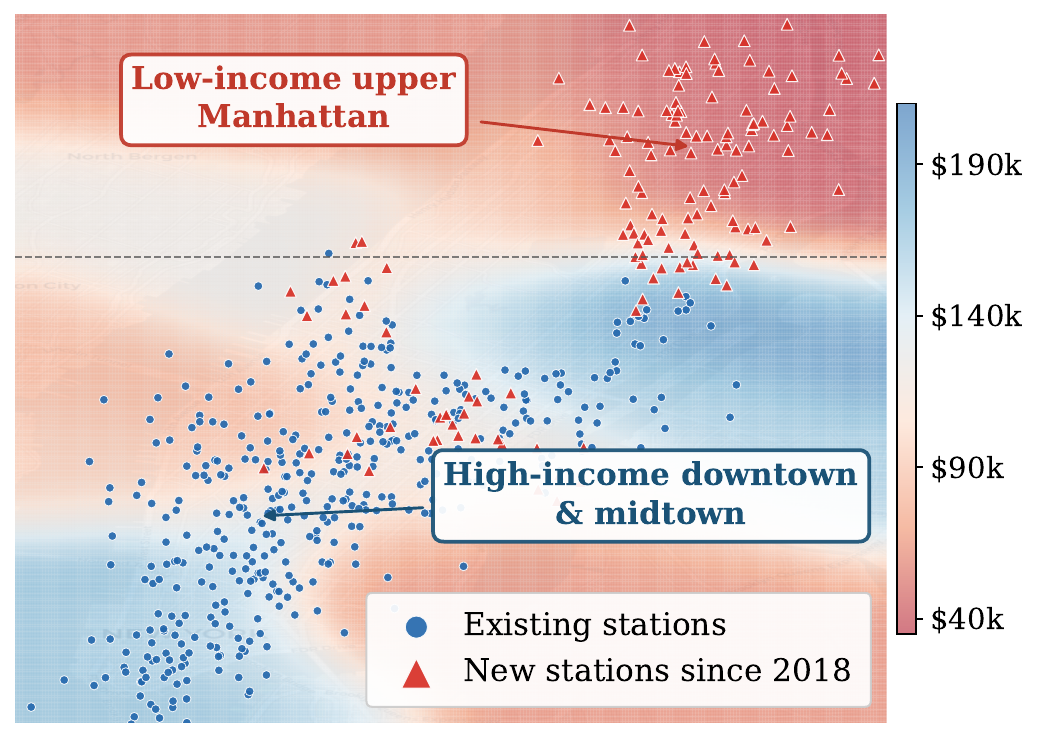}
        \subcaption{}
        \label{fig:subfig_a1}
    \end{minipage}
    \hfill
    \begin{minipage}[b]{.24\textwidth}
        \centering
        \includegraphics[width=\linewidth]{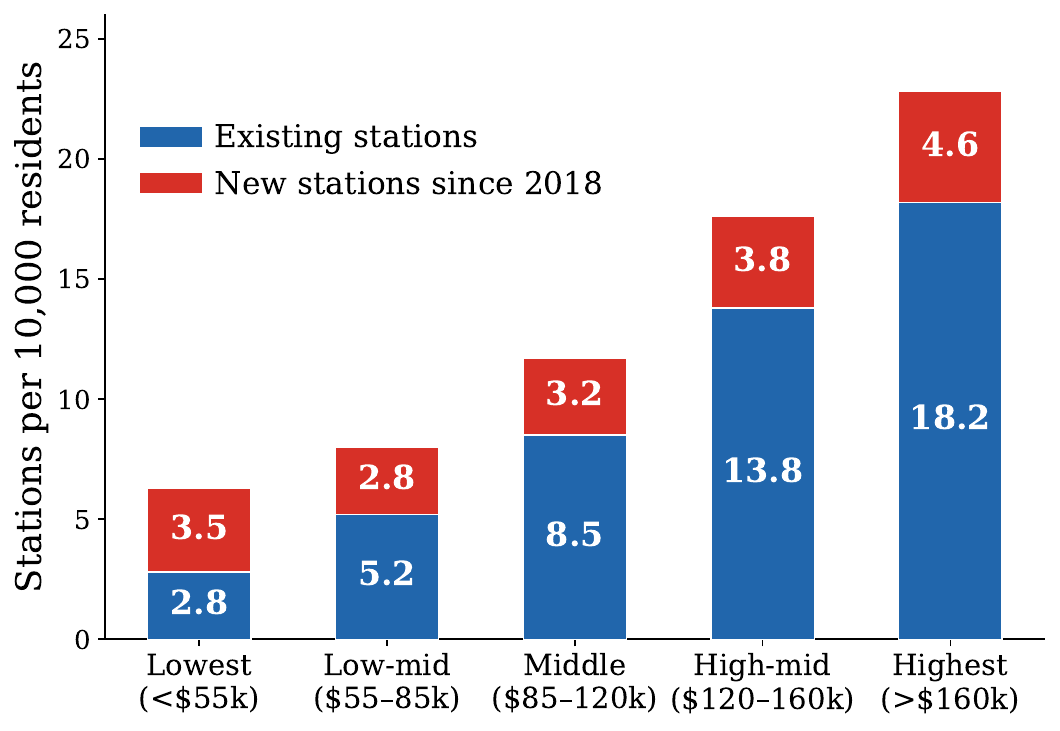}
        \subcaption{}
        \label{fig:subfig_b1}
    \end{minipage}
    \caption{Bike-sharing station coverage and neighbourhood income in Manhattan. (a) Spatial distribution of existing stations in 2018 and newly deployed stations from 2018. (b) Per-capita station density across neighbourhood income quintiles.}
    \label{fig:my_label1}
\end{figure}

Fairness has become an important concern in the expansion of low-carbon urban mobility systems. Bike-sharing systems, as a prominent example, provide short-term bicycle rental services through networks of spatially distributed stations and offer environmentally friendly alternatives for last-mile commuting. By reducing reliance on motorized trips, they can help alleviate traffic congestion and support broader carbon reduction objectives in cities pursuing climate action~\cite{amatuni2020does}. As illustrated in Fig.~\ref{fig:my_label1}(\subref{fig:subfig_a1}), Citi Bike stations in Manhattan have expanded substantially from 2018 to 2024~\cite{mahajan2024global}. However, this expansion remains spatially uneven, with newly deployed stations concentrated in high-income downtown and midtown corridors, while lower-income neighborhoods in Upper Manhattan remain comparatively underserved.

This uneven expansion raises a critical equity concern. As quantified in Fig.~\ref{fig:my_label1}(\subref{fig:subfig_b1}), the highest-income quintile (\(>\$160k\)) is served by 22.8 stations per 10,000 residents, more than three times the 6.3 stations available to the lowest-income quintile (\(<\$55k\)), indicating that recent deployments have widened the coverage gap. This pattern is consistent with evidence from North American cities showing that higher-income and more educated residents have disproportionately greater access to bike-sharing infrastructure~\cite{beaudoin2015public,venter2018equity}. Crucially, low observed ridership in underserved communities should not be treated as weak latent demand, as it may reflect affordability constraints, lower technology adoption, and the absence of nearby stations~\cite{wang2021data}. Without targeted intervention, such imbalance may reinforce green mobility gentrification, concentrating the benefits of low-carbon transportation infrastructure in already advantaged communities. This concern is compounded by a self-reinforcing dynamic inherent to data-driven deployment: stations in underserved communities generate lower observed ridership, causing predictive models to underestimate their true latent need, which further reduces their likelihood of being selected in demand-driven expansion planning~\cite{mehrabi2021survey, ensign2018runaway}. The result is a feedback loop in which algorithmic predictions reproduce and amplify the infrastructure gaps they are built to inform, a pattern that is particularly consequential when model outputs directly shape capital investment decisions in expanding networks.

Addressing this equity challenge requires accurate demand prediction for newly deployed stations, essential for guiding fair and efficient resource allocation during network expansion. However, new stations pose a fundamental modeling challenge because they lack historical ridership records and are embedded in a continuously evolving spatial network. Traditional regression-based models rely on historical demand observations to capture temporal regularities, but have limited capacity to model inter-station spatial dependencies~\cite{chen2015bike, liu2015station}. Functional zone-based approaches incorporate urban context, such as land use and population density, to support localized analysis~\cite{liu2017functional}, while machine learning models, including random forests and support vector machines, improve prediction accuracy through multi-source feature integration~\cite{kou2021incorporating}. More recently, graph neural network based approaches, such as STGCN~\cite{10.5555/3304222.3304273} and Graph WaveNet~\cite{wu2019graph}, have achieved strong performance in spatio-temporal prediction by modeling complex spatial dependencies among stations. Nevertheless, most methods are developed under a transductive setting with a fixed training graph, limiting their generalization to newly deployed stations. The challenge extends beyond topology evolution because newly deployed stations possess only location-based spatial attributes at inference time and lack the historical demand sequences required by graph-based models for reliable prediction. Recent transformer-based spatiotemporal architectures~\cite{jin2023spatio} relax the rigid structural assumptions of graph convolution, but they still depend on historical demand signals during inference. Consequently, these models remain subject to the same cold-start limitation when applied to stations that were absent from the training graph.

Beyond the prediction gap, this feature-level asymmetry between existing and newly deployed stations carries important fairness consequences. When a model encounters a new station at inference time with only spatial context available, its demand estimates are shaped by representations learned from existing stations whose patterns reflect historical inequities in infrastructure access. Areas with established transit connections, dense amenities, and high surrounding mobility tend to produce strong spatial signals, and are therefore more likely to receive high predicted demand regardless of the socioeconomic profile of the target location. Underserved communities with sparser surrounding infrastructure may receive systematically lower estimates even when their true latent demand is comparable. Consequently, resolving the cold-start prediction problem without simultaneously accounting for income-based disparities risks translating historical deployment inequities directly into demand predictions for newly deployed stations.

Motivated by the need to generalize beyond fixed training graphs, inductive methods such as GraphSAGE~\cite{liu2020graphsage}, KITS~\cite{xu2025kits}, and DA-MRGNN~\cite{liang2023cross} enable prediction for unseen nodes through neighborhood aggregation, knowledge transfer, and domain adaptation. Although these approaches improve generalization under evolving network structures, they remain primarily accuracy-driven and do not explicitly address income-based disparities in demand estimation. Fairness-aware methods such as FairST~\cite{yan2020fairness} introduce spatiotemporal fairness metrics and regularization, but are designed mainly for fixed grid-based representations and are not directly applicable to newly deployed stations absent during training. Foundation models such as UrbanGPT~\cite{li2024urbangpt} use large language models to infer spatiotemporal patterns from textual station descriptions, providing some zero-shot capability for new stations. However, they lack income-stratified supervision and cannot reliably correct systematic group-level prediction disparities through language-based spatial reasoning alone. Existing research therefore treats inductive demand prediction and fairness-aware mobility modeling largely as separate problems, leaving a methodological gap in jointly modeling evolving station-level graphs, demand at newly added stations, and income-stratified fairness in expanding bike-sharing systems.

To tackle these challenges, we propose \textbf{FairGIN}, a fairness-aware dynamic graph neural network framework for demand prediction in expanding bike-sharing systems. FairGIN is designed to jointly address two intertwined issues that arise during network expansion: the evolving topology of station-level graphs and the equity risks embedded in data-driven deployment decisions. To improve generalization to newly deployed stations, FairGIN introduces an Expansion-Simulated Increment Training strategy that treats selected existing stations as pseudo-new nodes during training. By exposing the model to expansion-like scenarios before deployment, this strategy reduces the discrepancy between the fully observed training graph and the partially observed graph encountered when new stations are introduced. For cold-start station representation, FairGIN further develops an Attention-Based Knowledge Transfer mechanism that adaptively transfers information from data-rich existing stations to data-sparse new stations through learnable soft attention, avoiding the rigidity of fixed neighbor selection. Beyond predictive accuracy, FairGIN incorporates a Fairness-Aware Training Objective that aligns demand forecasting with equity-oriented planning goals. This objective reduces income-based disparities in prediction, particularly for communities that have historically received limited access to bike-sharing infrastructure. 

The contributions of our work are summarized as follows:

\begin{itemize}
    \item We propose Expansion-Simulated Increment Training to bridge the training-inference graph gap by exposing the model to expansion-like scenarios, thereby enhancing generalization to newly deployed stations.

    \item We develop an Attention-Based Knowledge Transfer mechanism for adaptive cold-start representation learning, enabling new stations to selectively leverage information from data-rich existing stations.

    \item We incorporate income-stratified fairness regularization into the learning objective, reducing systematic prediction disparities and supporting equitable station deployment.

    \item  We validate our proposed approach using datasets from NYC and Seattle mobility systems. Experimental results consistently demonstrate superior performance compared to state-of-the-art baselines.
\end{itemize}
\section{Problem Statement}

This section formalizes the problem addressed by FairGIN, namely fairness-aware demand prediction for newly deployed bike-sharing stations under dynamic network expansion. We first define the station graph and feature representations, then introduce the prediction task, fairness formulation, joint optimization objective, and deployment scoring mechanism.

\subsection{Expanding Bike-Sharing Network Formulation}

Let $\mathcal{V}_A=\{v_1,v_2,\ldots,v_N\}$ denote the set of $N$ existing bike-sharing stations, and let $\mathcal{V}_B=\{u_1,u_2,\ldots,u_M\}$ denote the set of $M$ newly deployed stations. Each station is described by two categories of features. Location-based features $\mathbf{x}_i^{\mathrm{sp}}\in\mathbb{R}^{d_s}$ are derived from external data sources for any geographic location and include POI category distributions, road network characteristics, weather conditions, and ambient taxi flow as a proxy for surrounding mobility demand. These features are available for both existing and new stations. History-based features $\mathbf{x}_i^{\mathrm{tp}}\in\mathbb{R}^{T\times d_t}$ consist of historical hourly demand sequences over $T$ time steps and are available only for existing stations $v_i\in\mathcal{V}_A$. For newly deployed stations $u_j\in\mathcal{V}_B$, no ridership history exists, so $\mathbf{x}_j^{\mathrm{tp}}=\mathbf{0}$. 

We represent the expanding system as a dynamic graph $\mathcal{G}^{t} = (\mathcal{V}^{t}, \mathcal{E}^{t}, \mathbf{A}^{t})$ , where $\mathcal{V}=\mathcal{V}_A\cup\mathcal{V}_B$. The edge weight $a_{ij}^{t}$ is defined according to station availability:

\begin{equation}
\label{edge}
a_{ij}^{t} =
\begin{cases}
\begin{aligned}
&\alpha\,\mathrm{sim}_{sp}(i,j)
+ \beta\,\mathrm{dist}(i,j) \\
&\quad + \gamma\,\mathrm{sim}_{tp}(i,j),
\end{aligned}
& i,j \in \mathcal{V}_A, \\[6pt]
\begin{aligned}
&\alpha\,\mathrm{sim}_{sp}(i,j)
+ \beta\,\mathrm{dist}(i,j),
\end{aligned}
& i \in \mathcal{V}_A,\ j \in \mathcal{V}_B .
\end{cases}
\end{equation}

Here, $\mathrm{sim}_{sp}$, $\mathrm{dist}$, and $\mathrm{sim}_{tp}$ denote spatial feature similarity, geographic distance decay, and temporal demand correlation, respectively. The temporal term is omitted for edges involving newly deployed stations because these stations have no historical demand observations.

\subsection{Bike-Sharing Demand Prediction Formulation}

Given existing-station features $\{\mathbf{x}_i\}_{i \in \mathcal{V}_A}$, new-station spatial features $\{\mathbf{x}_j^{sp}\}_{j \in \mathcal{V}_B}$, and graph $\mathcal{G}$, the task is to learn a mapping $f_\theta$ that predicts hourly demand for new stations:

\begin{equation}
\hat{\mathbf{y}}_B
=
f_\theta\!\left(\mathcal{G}, \{\mathbf{x}_i\}_{i \in \mathcal{V}_A}, \{\mathbf{x}_j^{sp}\}_{j \in \mathcal{V}_B}\right),
\quad
\hat{\mathbf{y}}_B \in \mathbb{R}^M .
\end{equation}

The main challenge is that stations in $\mathcal{V}_B$ are unseen during training, so their demand must be inferred from spatial context and knowledge transferred from existing stations.

\subsection{Fairness Formulation}

\begin{figure*}[t] 
  \centering 
  \includegraphics[width=\textwidth]{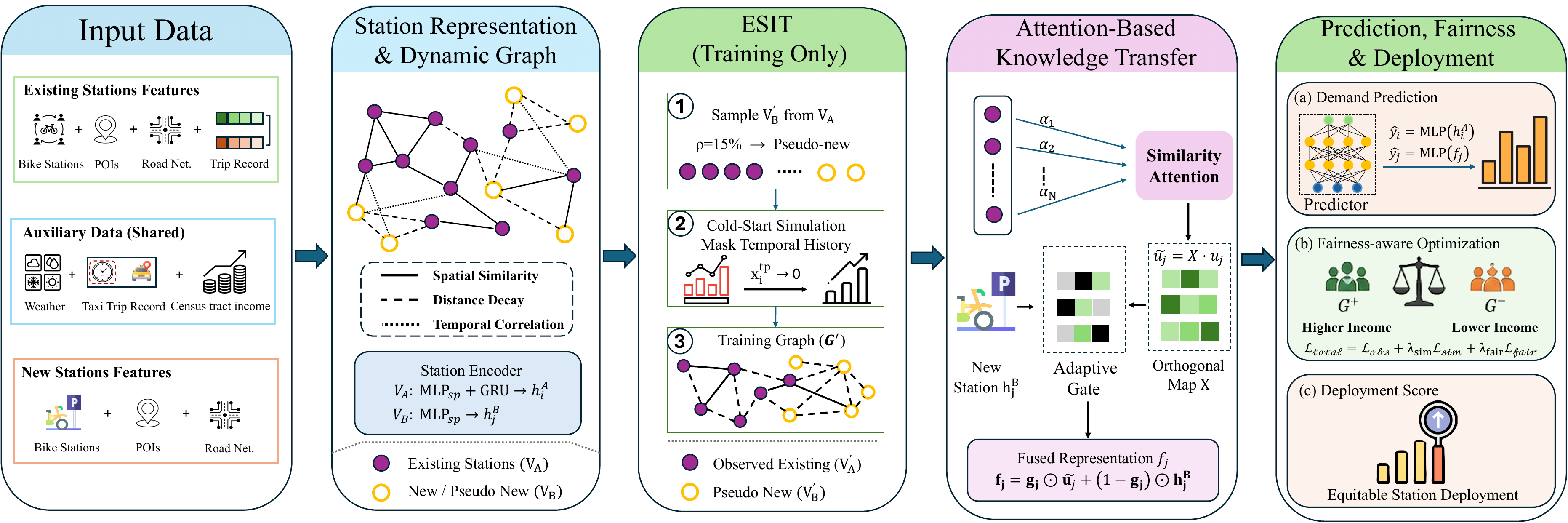} 
  \caption{Overview of the proposed FairGIN framework for fairness-aware demand prediction in expanding bike-sharing networks. The framework first constructs a heterogeneous station graph from multi-source station features, then applies expansion-simulated increment training (ESIT) to simulate cold-start expansion scenarios, and finally performs attention-based knowledge transfer with fairness-aware optimization for new-station demand prediction.}
  \label{fig:my_label2}
\end{figure*}

\textbf{Income group partition.}~We use median household income as the sensitive attribute, obtained from American Community Survey (ACS) 5-year estimates at the census tract level. Each candidate station $u_j \in \mathcal{V}_B$ is assigned to its enclosing census tract. Stations located in tracts at or above the city-wide median income form the advantaged group $\mathcal{G}^+$, while those below the median form the disadvantaged group $\mathcal{G}^-$.

\textbf{Fairness metrics.}~We evaluate prediction equity using two complementary metrics. The Region-based Fairness Gap~\cite{yan2020fairness} measures the per-capita predicted demand difference between income groups across all evaluated regions:

\begin{equation}
\mathrm{RFG} =
\left|
\frac{\sum_{i \in \mathcal{G}^+} \hat{y}_i}
     {\sum_{i \in \mathcal{G}^+} p_i}
-
\frac{\sum_{j \in \mathcal{G}^-} \hat{y}_j}
     {\sum_{j \in \mathcal{G}^-} p_j}
\right|,
\end{equation}
where $p_i$ denotes the population of station $i$'s service area normalized by the city total. A smaller RFG indicates lower group-level prediction disparity.

The Individual-based Fairness Gap~\cite{yan2020fairness} measures station-level disparity by representing each station as a weighted mixture of income groups:

\begin{equation}
\mathrm{IFG} =
\frac{
\sum_{i \in \mathcal{V}_B} w_i^{+} \hat{y}_i / p_i
}{
\sum_{i \in \mathcal{V}_B} w_i^{+}
}
-
\frac{
\sum_{i \in \mathcal{V}_B} w_i^{-} \hat{y}_i / p_i
}{
\sum_{i \in \mathcal{V}_B} w_i^{-}
},
\end{equation}
where $w_i^+$ and $w_i^-$ denote the proportions of residents in advantaged and disadvantaged income brackets within station $i$'s census tract. We also report Spearman's $\rho$ between predicted demand and tract-level income across $\mathcal{V}_B$, where $\rho \approx 0$ indicates weak income dependence.

The model is optimized by jointly minimizing prediction error and income-based demand disparity:

\begin{equation}
\mathcal{L}_{total}
=
\mathcal{L}_{obs}
+
\lambda_{sim}\mathcal{L}_{sim}
+
\lambda_{fair}\mathcal{L}_{fair},
\label{eq:objective}
\end{equation}

where $\mathcal{L}_{obs}$ supervises demand prediction on existing stations, $\mathcal{L}_{sim}$ supervises expansion-simulated pseudo-new stations, and $\mathcal{L}_{fair}$ penalizes per-capita demand disparity between $\mathcal{G}^+$ and $\mathcal{G}^-$. During inference, candidate stations are ranked by an income-aware deployment score $\mathrm{Score}_j = \hat{y}_j + \alpha \mathbf{1}[j \in \mathcal{G}^-]$, where $\alpha \geq 0$ controls the fairness bonus assigned to stations in $\mathcal{G}^-$ for deployment prioritization.
\section{Methodology}

This section presents the design of FairGIN for fairness-aware demand prediction at newly deployed stations in expanding bike-sharing systems. The overall framework, illustrated in Fig.~\ref{fig:my_label2}, comprises four tightly coupled components that jointly address dynamic network expansion, cold-start demand prediction, bias mitigation, and fairness-aware station deployment. The following subsections describe the design rationale and technical details of each component.

\subsection{Station Representation and Graph Construction}

Bike-sharing demand exhibits substantial spatial and temporal heterogeneity, while different station types provide different levels of observable information. To accommodate this heterogeneity, FairGIN constructs station representations according to the feature availability of each node type. For each existing station $v_i \in \mathcal{V}_A$, FairGIN encodes two complementary sources of information. Location-based spatial features $\mathbf{x}_i^{sp}$ (including POI distributions, road network characteristics, and taxi flow) are projected through a two-layer MLP, and temporal demand sequences $\mathbf{x}_i^{tp}$ are encoded by a GRU. The resulting representations are concatenated to obtain a unified station embedding $\mathbf{h}_i^A \in \mathbb{R}^d$. For newly deployed stations $u_j \in \mathcal{V}_B$, historical demand observations are unavailable; therefore, their initial representations are derived only from spatial features, i.e., $\mathbf{h}_j^B = \mathrm{MLP}_{sp}(\mathbf{x}_j^{sp}) \in \mathbb{R}^d$.

To model structural dependencies among stations, FairGIN constructs a dynamic graph $\mathcal{G} = (\mathcal{V}, \mathcal{E}, \mathbf{A})$ following Eq.~(\ref{edge}), with edge weights reflecting the different feature availability of existing and newly deployed stations. Contextual information is then propagated through $L$ graph convolution layers, where each layer updates station representations as $\mathbf{H}^{(l+1)} = \sigma(\hat{\mathbf{A}}\mathbf{H}^{(l)}\mathbf{W}^{(l)})$. Here, $\hat{\mathbf{A}}$ denotes the normalized adjacency matrix and $\mathbf{W}^{(l)}$ is the learnable transformation matrix at layer $l$. The resulting representations encode both station-level features and neighborhood structure for subsequent demand prediction tasks, especially for newly deployed stations with limited observations.

\subsection{Expansion-Simulated Increment Training}

A key challenge in predicting demand for newly deployed stations lies in the distribution mismatch between training and deployment. A model trained only on fully observed existing stations does not encounter nodes with missing temporal histories during training. When newly deployed stations appear at inference time, the model faces a different feature distribution, which can lead to degraded prediction performance.

To reduce this mismatch, we propose Expansion-Simulated Increment Training (ESIT), which simulates network expansion during each training epoch. At the beginning of each epoch, $\rho = 15\%$ of existing stations in $\mathcal{V}_A$ are uniformly sampled as pseudo-new stations, denoted as $\mathcal{V}_B' \subset \mathcal{V}_A$. Their historical demand sequences are masked as $\mathbf{x}_i^{tp} \leftarrow \mathbf{0}$ for all $i \in \mathcal{V}_B'$, reproducing the cold-start condition of actual newly deployed stations while preserving location-based features.

The remaining stations form the observed set $\mathcal{V}_A' = \mathcal{V}_A \setminus \mathcal{V}_B'$. The resulting training graph $\mathcal{G}' = (\mathcal{V}_A' \cup \mathcal{V}_B',\, \mathcal{E}')$ follows the same structural setting as the deployment graph, where observed stations coexist with new stations that lack historical demand records.

Since pseudo-new stations are sampled from existing stations, their ground-truth demand $y_i$ remains available and can be used to supervise the simulation loss. Their geographic coordinates also enable census-tract spatial matching to assign income group labels $\mathcal{G}^+$ and $\mathcal{G}^-$, allowing the fairness loss to be optimized during training without additional annotations. ESIT therefore improves generalization to cold-start stations while supporting income-stratified fairness supervision.

\subsection{Attention-Based Knowledge Transfer}

After graph message passing, newly deployed stations still lack historical demand information and require prediction-ready representations. FairGIN addresses this by using a differentiable attention-based transfer mechanism that aggregates knowledge from observed existing stations according to representation similarity. A learnable station-specific temperature further adapts the concentration of transferred knowledge to each new station.

For each pseudo-new station $j \in \mathcal{V}_B'$, similarity scores with observed existing stations are computed as $S_{ji} = \mathrm{sim}(\mathbf{h}_j^B, \mathbf{h}_i^A)$. The attention distribution is controlled by a station-specific temperature $\tau_j = \mathrm{softplus}(\mathbf{w}^\top \mathbf{h}_j^B + b)$, which is learned from the station embedding:
\begin{equation}
\alpha_{ji} =
\frac{\exp(S_{ji}/\tau_j)}
{\sum_{k \in \mathcal{V}_A'} \exp(S_{jk}/\tau_j)} .
\label{eq:attention}
\end{equation}
A smaller $\tau_j$ concentrates attention on the most similar existing stations, while a larger $\tau_j$ encourages broader aggregation. The transferred embedding is computed as $\mathbf{u}_j = \sum_{i \in \mathcal{V}_A'} \alpha_{ji}\mathbf{h}_i^A$.

To reduce distributional mismatch between existing and newly deployed station embeddings, the transferred representation is further transformed by an orthogonal mapping, $\tilde{\mathbf{u}}_j = \mathbf{X}\mathbf{u}_j$, where $\mathbf{X}$ is parameterized through a Cayley transform to preserve orthogonality during training. The transformed transfer signal is then combined with the station's own spatial representation through a learnable gate:

\begin{align}
\mathbf{g}_j &= \sigma\!\left(\mathbf{W}_g[\mathbf{h}_j^B;\tilde{\mathbf{u}}_j] + \mathbf{b}_g\right), \\
\mathbf{f}_j &= \mathbf{g}_j \odot \tilde{\mathbf{u}}_j + (1 - \mathbf{g}_j) \odot \mathbf{h}_j^B .
\end{align}

This gating mechanism allows FairGIN to balance transferred knowledge with local spatial information according to each station's context. A shared MLP decoder is then used for demand prediction, with $\hat{y}_i = \mathrm{MLP}(\mathbf{h}_i^A)$ for observed stations $i \in \mathcal{V}_A'$ and $\hat{y}_j = \mathrm{MLP}(\mathbf{f}_j)$ for pseudo-new stations $j \in \mathcal{V}_B'$.

\subsection{Fairness-Aware Optimization}

\textbf{Training objective.}~Following the joint optimization in Eq.~(\ref{eq:objective}), FairGIN is trained with three loss terms. The observation loss preserves prediction accuracy on fully observed existing stations:

\begin{equation}
\mathcal{L}_{obs} =
\frac{1}{|\mathcal{V}_A'|}
\sum_{i \in \mathcal{V}_A'}(\hat{y}_i - y_i)^2 .
\end{equation}

The simulation loss provides direct supervision for pseudo-new stations generated by ESIT:
\begin{equation}
\mathcal{L}_{sim} =
\frac{1}{|\mathcal{V}_B'|}
\sum_{j \in \mathcal{V}_B'}(\hat{y}_j - y_j)^2 .
\end{equation}

The fairness loss penalizes income-based per-capita prediction disparity among pseudo-new stations:
\begin{equation}
\mathcal{L}_{fair} =
\left|
\frac{1}{|\mathcal{G}^+|}
\sum_{i \in \mathcal{G}^+}
\frac{\hat{y}_i}{p_i}
-
\frac{1}{|\mathcal{G}^-|}
\sum_{j \in \mathcal{G}^-}
\frac{\hat{y}_j}{p_j}
\right| .
\end{equation}

Here, $p_i$ denotes the population-normalized service area of station $i$. The hyperparameters $\lambda_{sim}$ and $\lambda_{fair}$ (in Eq.~\eqref{eq:objective}) are selected through grid search on the validation set.

\textbf{Demand prediction for new stations.}~During inference, ESIT is disabled and FairGIN is applied to the expanded graph $\mathcal{G} = (\mathcal{V}_A \cup \mathcal{V}_B, \mathcal{E})$. For each candidate new station $j \in \mathcal{V}_B$, only deployment-time spatial features $\mathbf{x}_j^{sp}$ are available, whereas historical demand observations remain unavailable. The trained attention-gate pipeline computes similarity scores $S_{ji}$ between the candidate station and existing station embeddings $\{\mathbf{h}_i^A\}$, generates the fused representation $\mathbf{f}_j$ by adaptively aggregating knowledge from relevant existing stations, and predicts demand through the shared decoder as $\hat{y}_j = \mathrm{MLP}(\mathbf{f}_j)$. Since ESIT explicitly aligns the training process with this inference condition, FairGIN can be directly applied to newly deployed stations without retraining, additional graph reconstruction, or manual feature imputation.

\textbf{Fairness-guided expansion.}~At inference time, FairGIN predicts demand $\hat{y}_j$ for each candidate new station $j \in \mathcal{V}_B$. Since ranking solely by predicted demand may still favor historically high-ridership areas, we apply the deployment score as a decision-level correction, where an equity bonus $\alpha$ is assigned to stations in $\mathcal{G}^-$. This score can be calibrated according to local equity objectives without modifying the trained model. In our experiments, we compare three deployment strategies: demand-only ranking by $\hat{y}_j$, random selection as an uninformed reference, and FairGIN ranking by $\mathrm{Score}_j$ with income-group adjustment. 
\section{Experiments}

In this section, we present the experimental datasets, implementation details, and evaluation protocol. We then compare FairGIN with state-of-the-art baselines, followed by ablation studies that assess the contribution of each proposed component. Sensitivity and robustness analyses are further conducted to examine the stability of key design choices. Finally, we simulate real-world deployment scenarios to evaluate the practical equity impact of FairGIN's deployment scoring mechanism.

\subsection{Experimental Setup}

\textbf{Datasets.}~We evaluate FairGIN on two publicly available bike-sharing datasets from major US cities with clear income variation across urban neighborhoods. Summary statistics for both datasets are reported in Table~\ref{tab:my_label1}.

NYC Citi Bike\footnote{\url{https://citibikenyc.com/system-data}} trip records are obtained from the NYC Open Data portal, covering January 2018 to December 2023. During this period, the system expanded from 746 to 1,973 docking stations. We treat stations active before January 2021 as the existing station set $\mathcal{V}_A$ with $N = 746$, and stations first activated between January 2021 and December 2023 as the new station evaluation set $\mathcal{V}_B$ with $M = 312$. Hourly demand is aggregated from more than 80 million trip records. The income variation across NYC neighborhoods provides a suitable setting for evaluating fairness-aware demand prediction.

Seattle Bikeshare trip records are obtained from the Transportation Data Collaborative at the University of Washington \cite{yan2020fairness}, covering October 2017 to October 2018 with more than 1.6 million trips across 247 stations. We apply a temporal split to define existing stations from the first eight months and new stations from the final four months, resulting in $N = 186$ existing stations and $M = 61$ new stations. Seattle provides a complementary evaluation setting due to its income variation between central and peripheral neighborhoods.

\begin{table}[t]
\centering
\caption{Dataset statistics.}
\label{tab:my_label1}
\setlength{\tabcolsep}{5pt}
\begin{tabular}{lcc}
\toprule
& \textbf{NYC Citi Bike} & \textbf{Seattle Bikeshare} \\
\midrule
Existing stations ($N$)     & 746          & 186         \\
New stations ($M$)          & 312          & 61          \\
Time span                   & 2018 to 2023 & 2017 to 2018 \\
Total trips                 & 80M+         & 1.6M+       \\
Temporal resolution         & Hourly       & Hourly      \\
G$^+$ / G$^-$ (new)         & 149 / 163    & 29 / 32     \\
City income median          & \$72,800     & \$93,500    \\
\bottomrule
\end{tabular}
\end{table}

\begin{table*}[t]
\centering
\caption{Prediction accuracy and fairness comparison on new stations ($\mathcal{V}_B$). \textbf{Bold}: best; \underline{underline}: best baseline per metric.}
\setlength{\tabcolsep}{5.2pt}
\resizebox{\textwidth}{!}{%
\begin{tabular}{lcccccccccc}
\toprule
& \multicolumn{5}{c}{\textbf{NYC Citi Bike}} & \multicolumn{5}{c}{\textbf{Seattle Bikeshare}} \\
\cmidrule(lr){2-6}\cmidrule(lr){7-11}
Method
  & MAE$\downarrow$ & RMSE$\downarrow$ & RFG$\downarrow$ & IFG$\downarrow$ & $|\rho|$$\downarrow$
  & MAE$\downarrow$ & RMSE$\downarrow$ & RFG$\downarrow$ & IFG$\downarrow$ & $|\rho|$$\downarrow$ \\
\midrule
ARIMA
  & 4.67$_{\pm.24}$ & 7.53$_{\pm.37}$ & 2.47$_{\pm.08}$ & 2.04$_{\pm.07}$ & 0.433$_{\pm.016}$
  & 3.89$_{\pm.21}$ & 6.24$_{\pm.32}$ & 2.08$_{\pm.07}$ & 1.72$_{\pm.06}$ & 0.414$_{\pm.015}$ \\
LSTM
  & 4.23$_{\pm.18}$ & 6.87$_{\pm.24}$ & 2.34$_{\pm.07}$ & 1.92$_{\pm.06}$ & 0.412$_{\pm.014}$
  & 3.47$_{\pm.14}$ & 5.61$_{\pm.19}$ & 1.98$_{\pm.06}$ & 1.67$_{\pm.05}$ & 0.387$_{\pm.013}$ \\
STGCN
  & 3.81$_{\pm.15}$ & 6.12$_{\pm.21}$ & 2.28$_{\pm.06}$ & 1.87$_{\pm.06}$ & 0.398$_{\pm.013}$
  & 3.12$_{\pm.12}$ & 5.08$_{\pm.18}$ & 1.93$_{\pm.06}$ & 1.62$_{\pm.05}$ & 0.371$_{\pm.012}$ \\
DCRNN
  & 3.64$_{\pm.14}$ & 5.93$_{\pm.20}$ & 2.31$_{\pm.07}$ & 1.89$_{\pm.06}$ & 0.403$_{\pm.013}$
  & 2.98$_{\pm.12}$ & 4.87$_{\pm.17}$ & 1.96$_{\pm.06}$ & 1.64$_{\pm.05}$ & 0.378$_{\pm.012}$ \\
FairST
  & 3.73$_{\pm.14}$ & 6.03$_{\pm.20}$ & \underline{1.71}$_{\pm.05}$ & \underline{1.42}$_{\pm.04}$ & \underline{0.315}$_{\pm.010}$
  & 3.06$_{\pm.12}$ & 4.95$_{\pm.17}$ & \underline{1.38}$_{\pm.04}$ & \underline{1.16}$_{\pm.04}$ & \underline{0.297}$_{\pm.010}$ \\
GraphSAGE
  & 3.17$_{\pm.13}$ & 5.24$_{\pm.18}$ & 2.09$_{\pm.06}$ & 1.74$_{\pm.05}$ & 0.363$_{\pm.012}$
  & 2.61$_{\pm.11}$ & 4.31$_{\pm.16}$ & 1.77$_{\pm.05}$ & 1.49$_{\pm.05}$ & 0.342$_{\pm.011}$ \\
DA-MRGNN
  & 2.78$_{\pm.11}$ & 4.56$_{\pm.16}$ & 1.97$_{\pm.05}$ & 1.63$_{\pm.05}$ & 0.341$_{\pm.011}$
  & 2.27$_{\pm.09}$ & 3.82$_{\pm.14}$ & 1.68$_{\pm.05}$ & 1.41$_{\pm.04}$ & 0.318$_{\pm.011}$ \\
KITS
  & \underline{2.65}$_{\pm.10}$ & \underline{4.38}$_{\pm.15}$ & 2.02$_{\pm.06}$ & 1.68$_{\pm.05}$ & 0.349$_{\pm.011}$
  & \underline{2.19}$_{\pm.09}$ & \underline{3.68}$_{\pm.13}$ & 1.71$_{\pm.05}$ & 1.44$_{\pm.04}$ & 0.326$_{\pm.011}$ \\
UrbanGPT
  & 2.91$_{\pm.12}$ & 4.74$_{\pm.16}$ & 1.89$_{\pm.05}$ & 1.57$_{\pm.05}$ & 0.327$_{\pm.010}$
  & 2.43$_{\pm.10}$ & 3.97$_{\pm.14}$ & 1.61$_{\pm.05}$ & 1.36$_{\pm.04}$ & 0.309$_{\pm.010}$ \\
\midrule
\textbf{FairGIN}
  & \textbf{2.23}$_{\pm.04}$ & \textbf{3.71}$_{\pm.07}$ & \textbf{1.14}$_{\pm.02}$ & \textbf{0.89}$_{\pm.02}$ & \textbf{0.187}$_{\pm.008}$
  & \textbf{1.87}$_{\pm.04}$ & \textbf{3.14}$_{\pm.06}$ & \textbf{0.93}$_{\pm.02}$ & \textbf{0.74}$_{\pm.02}$ & \textbf{0.162}$_{\pm.007}$ \\
\textit{Improv.}
  & \textit{15.8\%} & \textit{15.3\%} & \textit{33.3\%} & \textit{37.3\%} & \textit{40.6\%}
  & \textit{14.6\%} & \textit{14.7\%} & \textit{32.6\%} & \textit{36.2\%} & \textit{45.5\%} \\
\bottomrule
\end{tabular}%
}
\label{tab:my_label2}
\end{table*}

\textbf{Auxiliary data.}~To construct location-based features available for all stations, we incorporate publicly available POI distributions from OpenStreetMap\footnote{\url{https://www.openstreetmap.org}} within a 300\,m radius, covering food, retail, transit, parks, education, healthcare, entertainment, and offices. Road network~\footnote{\url{https://osmnx.readthedocs.io}} attributes are extracted using OSMnx. Weather features, including temperature, precipitation, and wind speed, are obtained from NOAA~\footnote{\url{https://www.ncei.noaa.gov}}. Ambient taxi flow within a 500\,m radius is derived from NYC TLC~\footnote{\url{https://www.nyc.gov/site/tlc/about/tlc-trip-record-data.page}} and Seattle Rideshare trip records to capture surrounding mobility activity. Income group labels ($\mathcal{G}^+$, $\mathcal{G}^-$) and income proportion weights ($w_i^+$, $w_i^-$) are derived from ACS\footnote{\url{https://data.census.gov}} 5-year estimates via census tract spatial matching, using Tables~B19013 and B19001, respectively.

\textbf{Implementation details.}~FairGIN is implemented in PyTorch 2.0 with PyTorch Geometric 2.3. The spatial encoder is a two-layer MLP with hidden dimension $d = 128$, and the temporal encoder is a single-layer GRU with hidden dimension 128. We use $L = 2$ graph convolutional layers. All models are optimized with Adam using a learning rate of $10^{-3}$ and weight decay of $10^{-4}$. Early stopping is applied with patience 20. The ESIT mask ratio is set to $\rho = 0.15$. The hyperparameters $\lambda_{sim} = 0.5$ and $\lambda_{fair} = 0.3$ are selected through grid search over $\{0.1, 0.3, 0.5, 0.8, 1.0\}^2$. The deployment equity bonus is set to $\alpha = 0.2$. All experiments are conducted on a single RTX4090, and results are reported as mean $\pm$ standard deviation. The model implementation is available at: \url{https://github.com/VineYX/FairGIN.git}.

\textbf{Compared methods.}~We compare FairGIN with nine representative baselines spanning statistical forecasting, sequential modeling, graph-based prediction, inductive learning, fairness-aware learning, and urban foundation models. All methods are trained on $\mathcal{V}_A$ and evaluated on $\mathcal{V}_B$ under the same protocol. For transductive models, the temporal features of new stations are unavailable and are therefore set to zero.

\begin{itemize}

\item \textbf{ARIMA}~\cite{box2015time}: A classical autoregressive integrated moving average model. For new stations, demand signals are first spatially interpolated from the $k$-nearest existing stations and then used for forecasting.

\item \textbf{LSTM}~\cite{hochreiter1997long}: A sequence-to-sequence recurrent model applied independently to each station. New-station temporal inputs are initialized to zero.

\item \textbf{STGCN}~\cite{10.5555/3304222.3304273}: A spatiotemporal graph convolutional network. New stations are inserted into the graph with zero temporal features and connected using the same distance-based adjacency as FairGIN.

\item \textbf{DCRNN}~\cite{li2017diffusion}: A diffusion convolutional recurrent network. It is adapted in the same way as STGCN by extending the graph topology and setting new-station temporal features to zero.

\item \textbf{FairST}~\cite{yan2020fairness}: A fairness-aware spatiotemporal forecasting baseline that incorporates income-stratified regularization into an STGCN backbone. The fairness constraint is applied to both existing and new stations.

\item \textbf{GraphSAGE}~(GSAGE)~\cite{hamilton2017inductive}: An inductive graph neural network that generalizes to unseen nodes through neighborhood aggregation. Spatial features are used as node attributes for new stations.

\item \textbf{DA-MRGNN}~(DA-MR)~\cite{liang2023cross}: A domain-adaptive multi-relational graph network for transfer, adapted by treating $\mathcal{V}_A$ as the source domain and $\mathcal{V}_B$ as the target domain.

\item \textbf{KITS}~\cite{xu2025kits}: A knowledge injection framework for cold-start traffic forecasting in expanding networks. It is applied directly, as it supports new-node prediction through knowledge transfer from existing nodes.

\item \textbf{UrbanGPT}~(UrGPT)~\cite{li2024urbangpt}: A large language model-based method for zero-shot urban spatio-temporal prediction. We provide detailed station metadata and local spatial context as textual prompts, while omitting historical temporal observations for newly deployed stations.

\end{itemize}

\textbf{Evaluation metrics.}~Prediction accuracy is evaluated using Mean Absolute Error (MAE) and Root Mean Squared Error (RMSE) on new stations $\mathcal{V}_B$. Fairness is measured by the Region-based Fairness Gap (RFG) and Individual-based Fairness Gap (IFG), where lower values indicate more equitable predictions. We also report the absolute Spearman rank correlation $|\rho|$ between predicted demand and census tract income, where values closer to zero indicate weaker dependence on income.

\subsection{Main Performance Comparison}

As shown in Table~\ref{tab:my_label2}, FairGIN achieves the lowest MAE and RMSE on both datasets, while also obtaining the smallest RFG, IFG, and Spearman $|\rho|$ among all compared methods. This joint improvement indicates that the gains in prediction accuracy are not achieved at the expense of fairness across income groups. On NYC Citi Bike, FairGIN reduces MAE by 15.8\% and RMSE by 15.3\% compared with the strongest accuracy baseline, KITS. It also reduces RFG by 33.3\% and IFG by 37.3\% compared with the strongest fairness baseline, FairST. Similar improvements are observed on Seattle Bikeshare, indicating that FairGIN generalizes across different city scales and system sizes in expansion settings. The consistently lower Spearman $|\rho|$ further suggests that its predicted demand is less strongly associated with neighborhood income levels.

The comparison reveals a clear and consistent pattern. Statistical and fixed-topology methods, including ARIMA, LSTM, STGCN, and DCRNN, struggle under the cold-start setting because they cannot effectively model newly deployed stations without historical demand observations. In particular, fixed-topology graph models are sensitive to the discrepancy between the training graph and the expanded inference graph. Inductive and cold-start-oriented methods, including GraphSAGE, DA-MRGNN, and KITS, improve prediction accuracy but provide limited fairness gains because they do not incorporate income-stratified supervision or explicitly constrain group-level prediction disparities. FairST achieves the strongest fairness performance among baselines through explicit fairness regularization, but its transductive STGCN backbone limits its accuracy on new stations under dynamic expansion. UrbanGPT shows competitive fairness but remains less accurate than specialized graph-based methods, highlighting the importance of explicitly modeling spatial dependencies among stations. FairGIN achieves the most balanced performance by jointly addressing expansion-induced cold-start prediction and income-aware fairness within a unified framework.

\begin{figure}[t]
    \centering
    \begin{minipage}[b]{.23\textwidth}
        \centering
        \includegraphics[width=\linewidth]{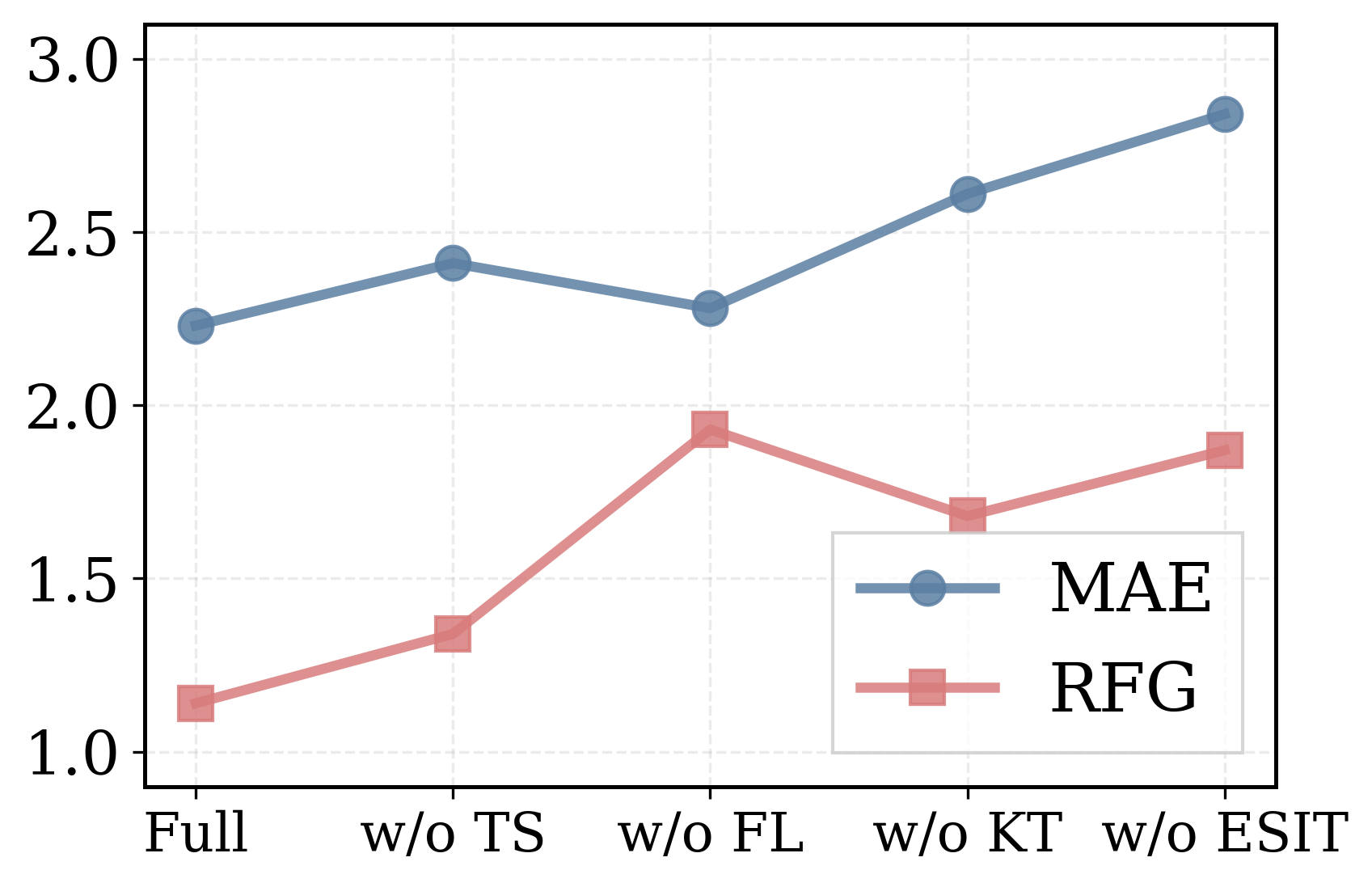}
        \subcaption{}
        \label{fig:subfig_a3}
    \end{minipage}
    \hfill
    \begin{minipage}[b]{.23\textwidth}
        \centering
        \includegraphics[width=\linewidth]{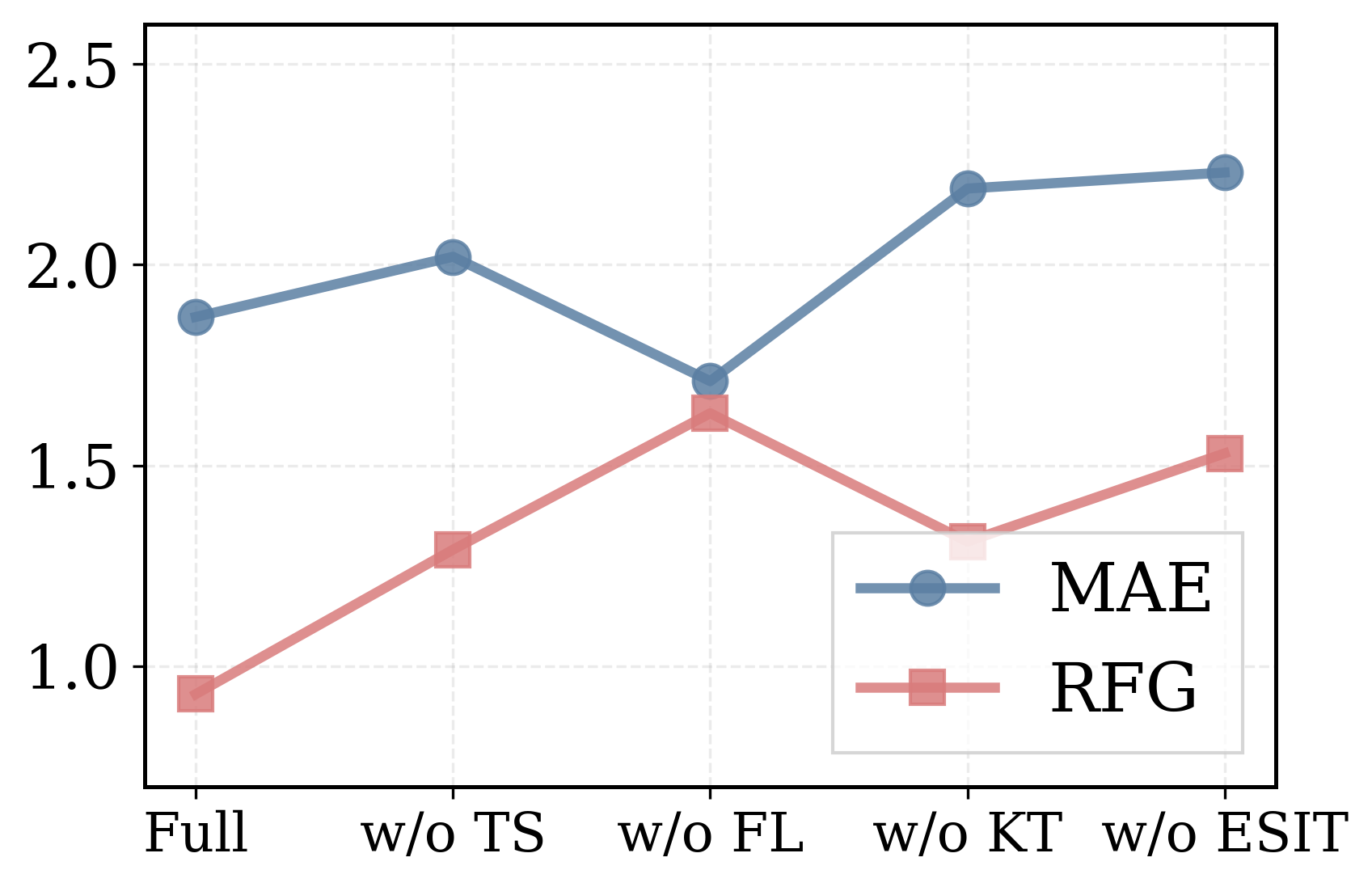}
        \subcaption{}
        \label{fig:subfig_b3}
    \end{minipage}
    \caption{Performance comparison of ablation variants evaluated by MAE and RFG. (a) NYC. (b) Seattle.}
    \label{fig:my_label3}
\end{figure}

\subsection{Ablation Study}

\begin{figure*}[t]
    \centering
    \begin{minipage}[b]{.235\textwidth}
        \centering
        \includegraphics[width=\linewidth]{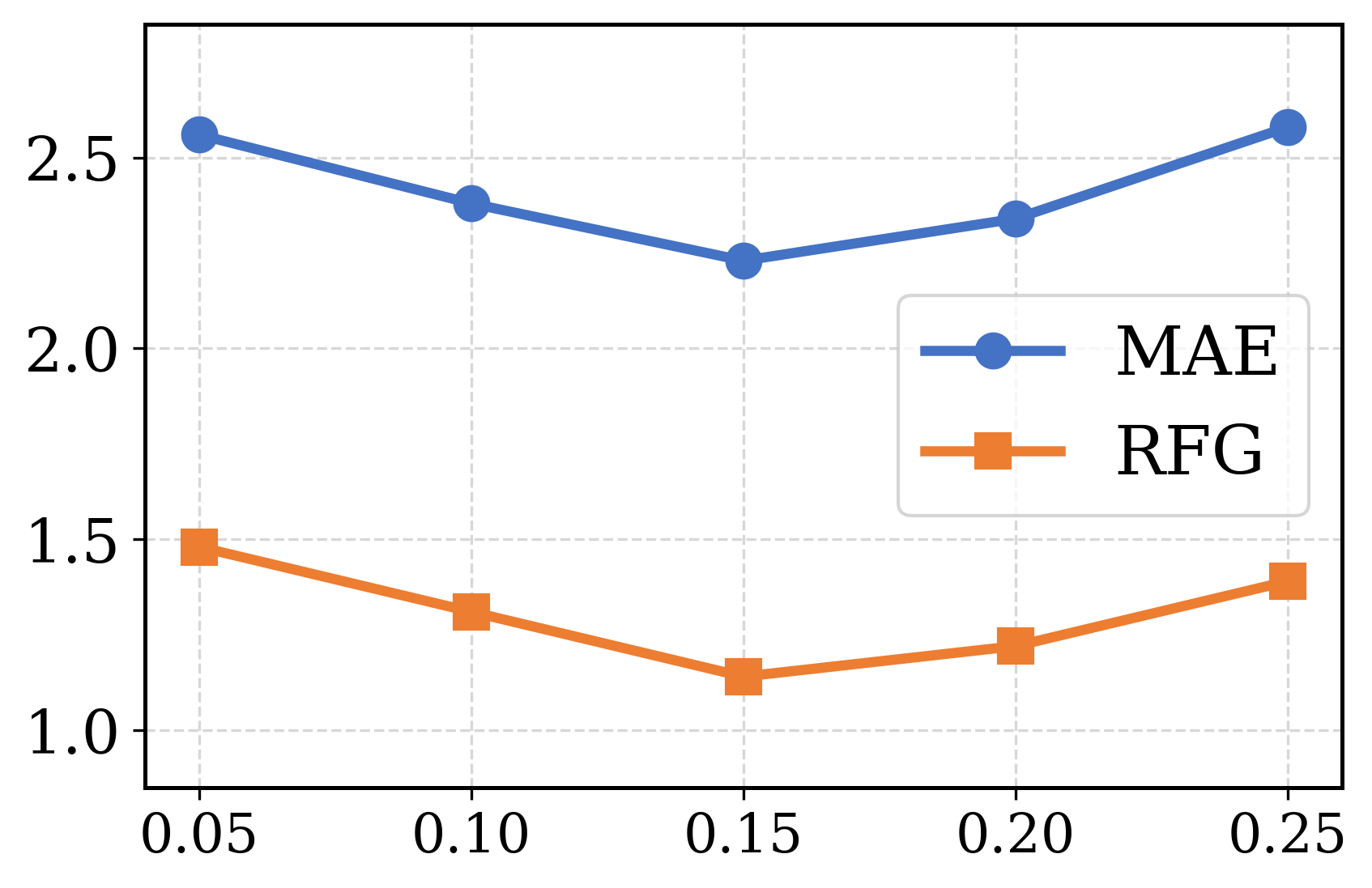}
        \subcaption{}
        \label{fig:subfig_a4}
    \end{minipage}
    \hfill
    \begin{minipage}[b]{.235\textwidth}
        \centering
        \includegraphics[width=\linewidth]{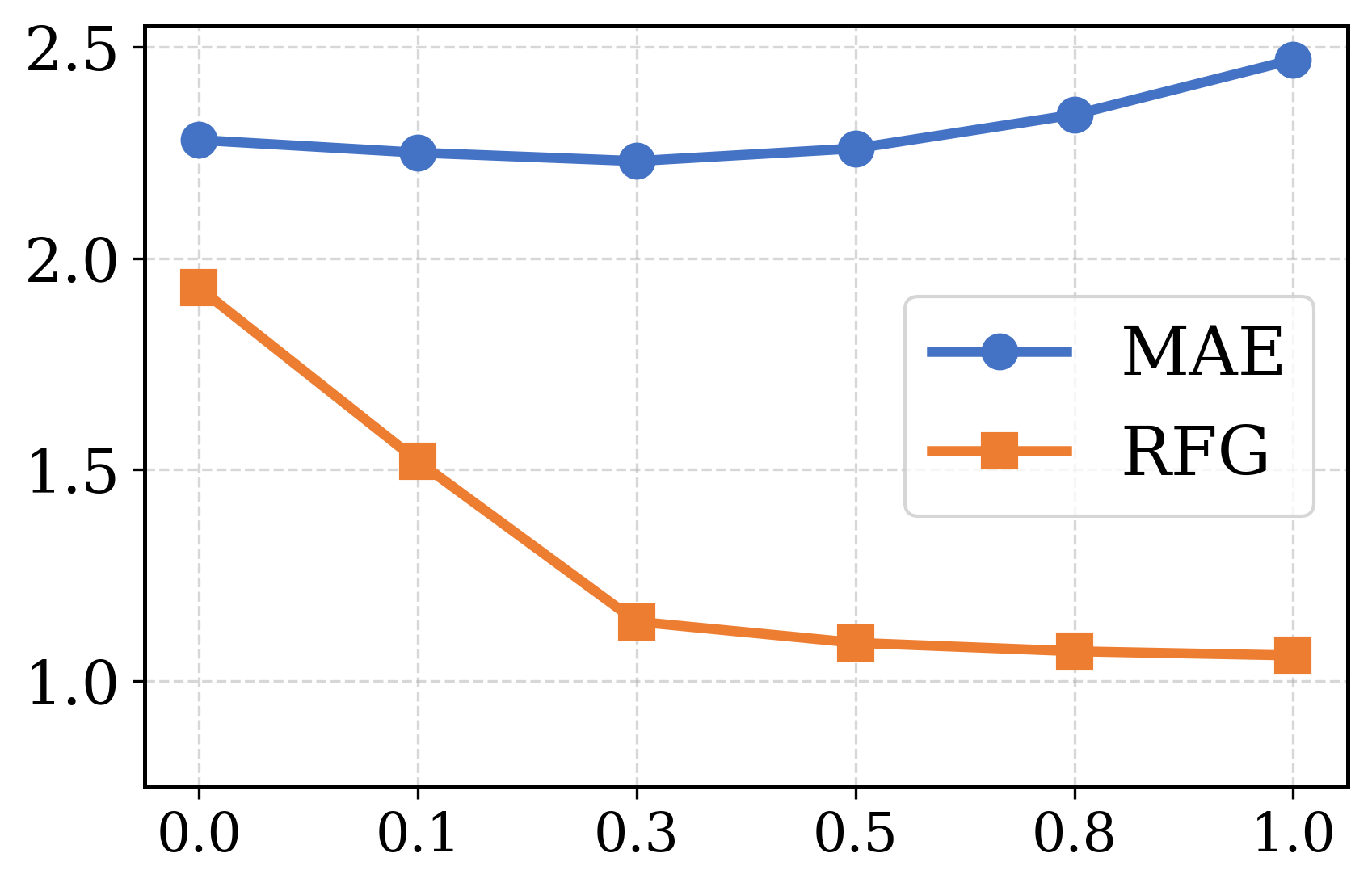}
        \subcaption{}
        \label{fig:subfig_b4}
    \end{minipage}
    \hfill
    \begin{minipage}[b]{.235\textwidth}
        \centering
        \includegraphics[width=\linewidth]{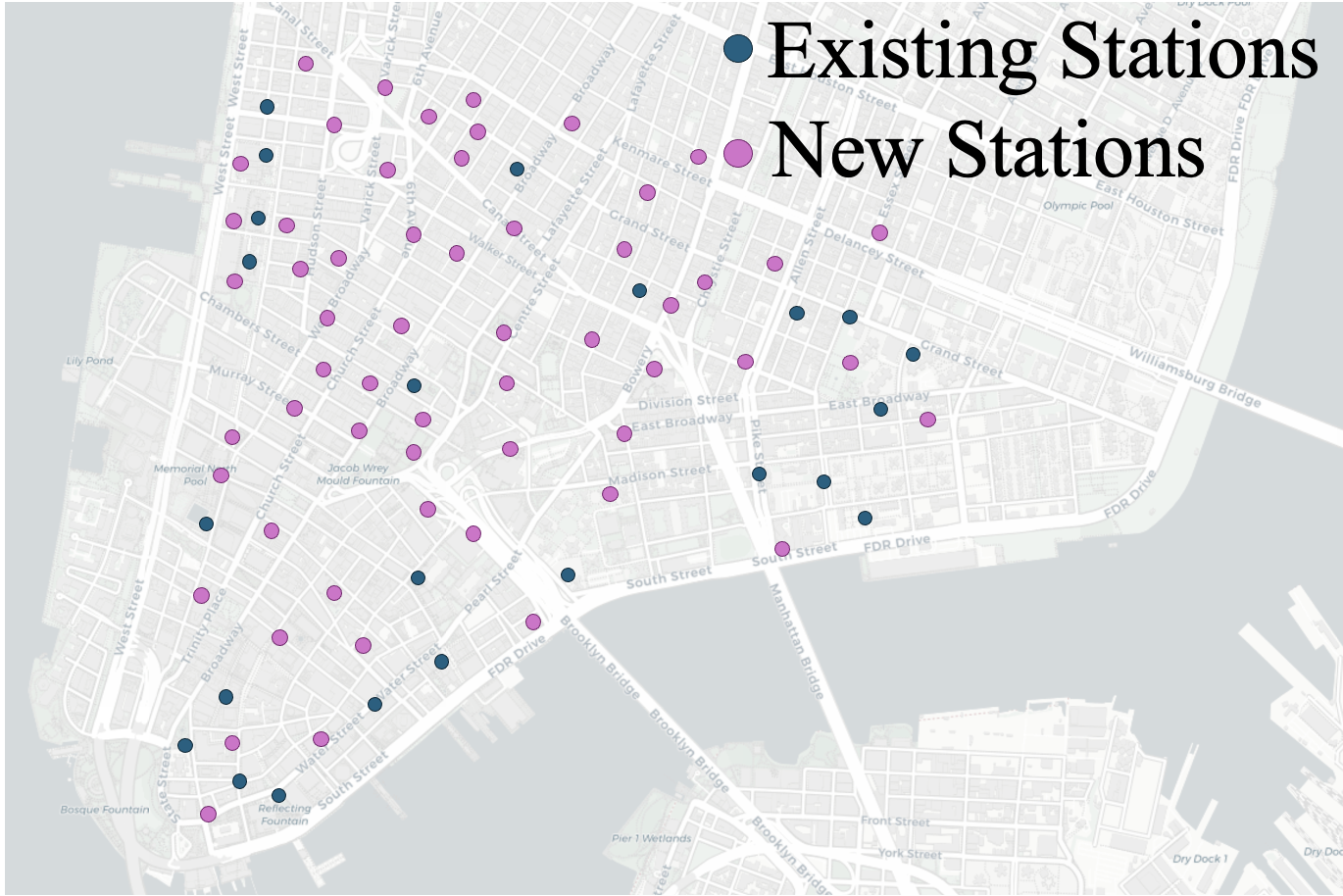}
        \subcaption{}
        \label{fig:subfig_c4}
    \end{minipage}
    \hfill
    \begin{minipage}[b]{.235\textwidth}
        \centering
        \includegraphics[width=\linewidth]{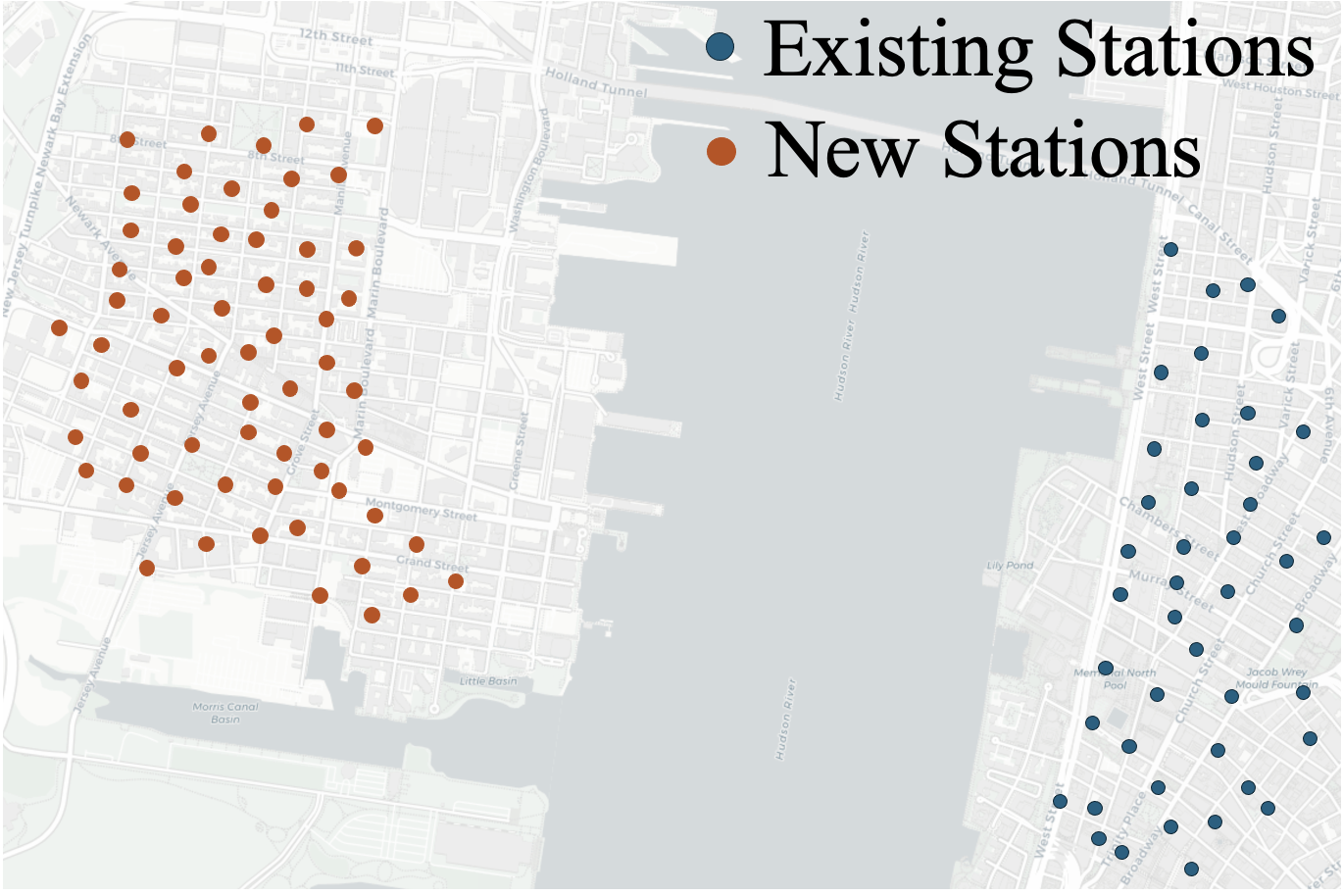}
        \subcaption{}
        \label{fig:subfig_a5}
    \end{minipage}

    \vspace{0.6em}

    \begin{minipage}[b]{.235\textwidth}
        \centering
        \includegraphics[width=\linewidth]{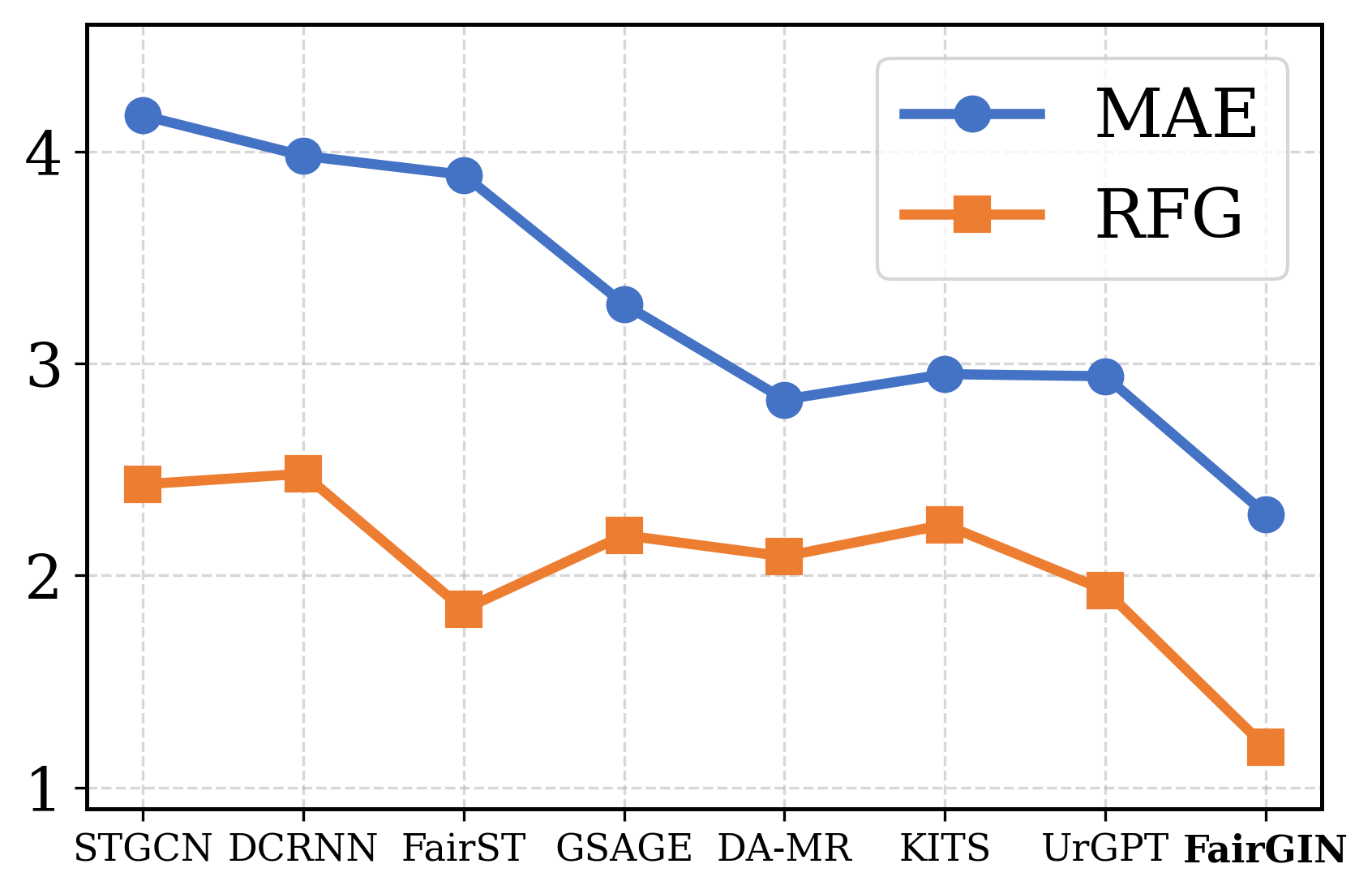}
        \subcaption{}
        \label{fig:subfig_a7}
    \end{minipage}
    \hfill
    \begin{minipage}[b]{.235\textwidth}
        \centering
        \includegraphics[width=\linewidth]{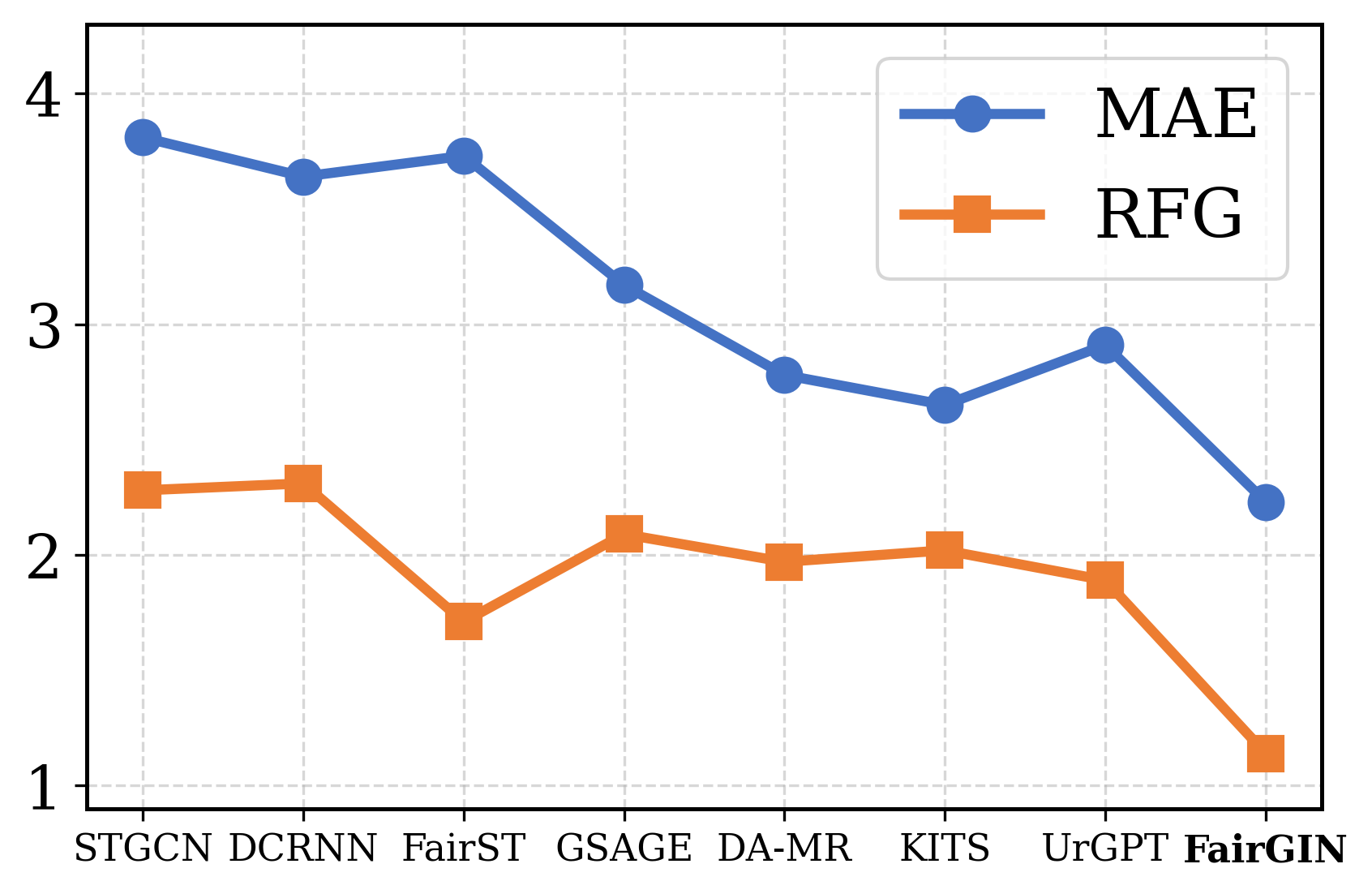}
        \subcaption{}
        \label{fig:subfig_a6}
    \end{minipage}
    \hfill
    \begin{minipage}[b]{.235\textwidth}
        \centering
        \includegraphics[width=\linewidth]{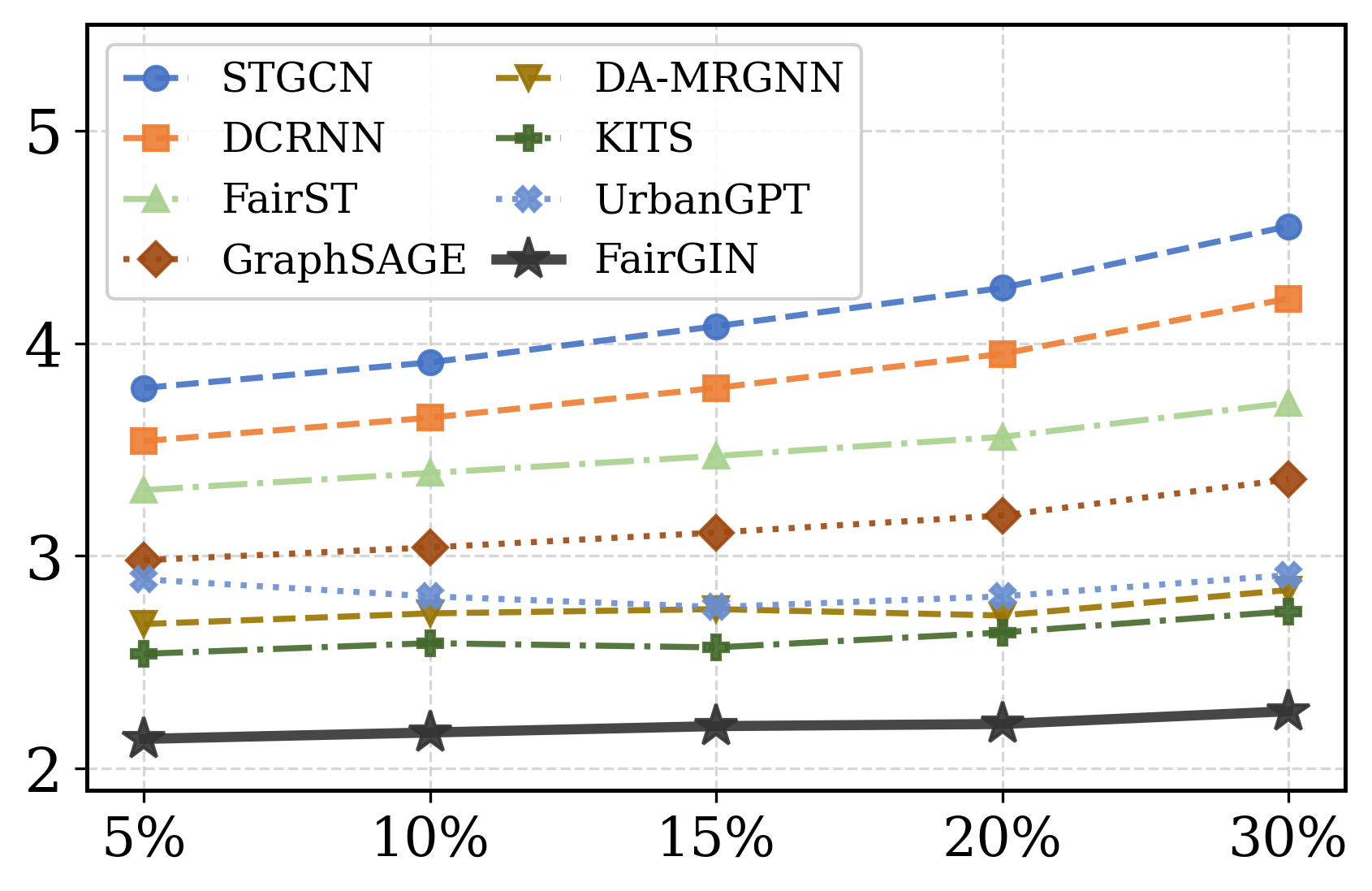}
        \subcaption{}
        \label{fig:subfig_b5}
    \end{minipage}
    \hfill
    \begin{minipage}[b]{.235\textwidth}
        \centering
        \includegraphics[width=\linewidth]{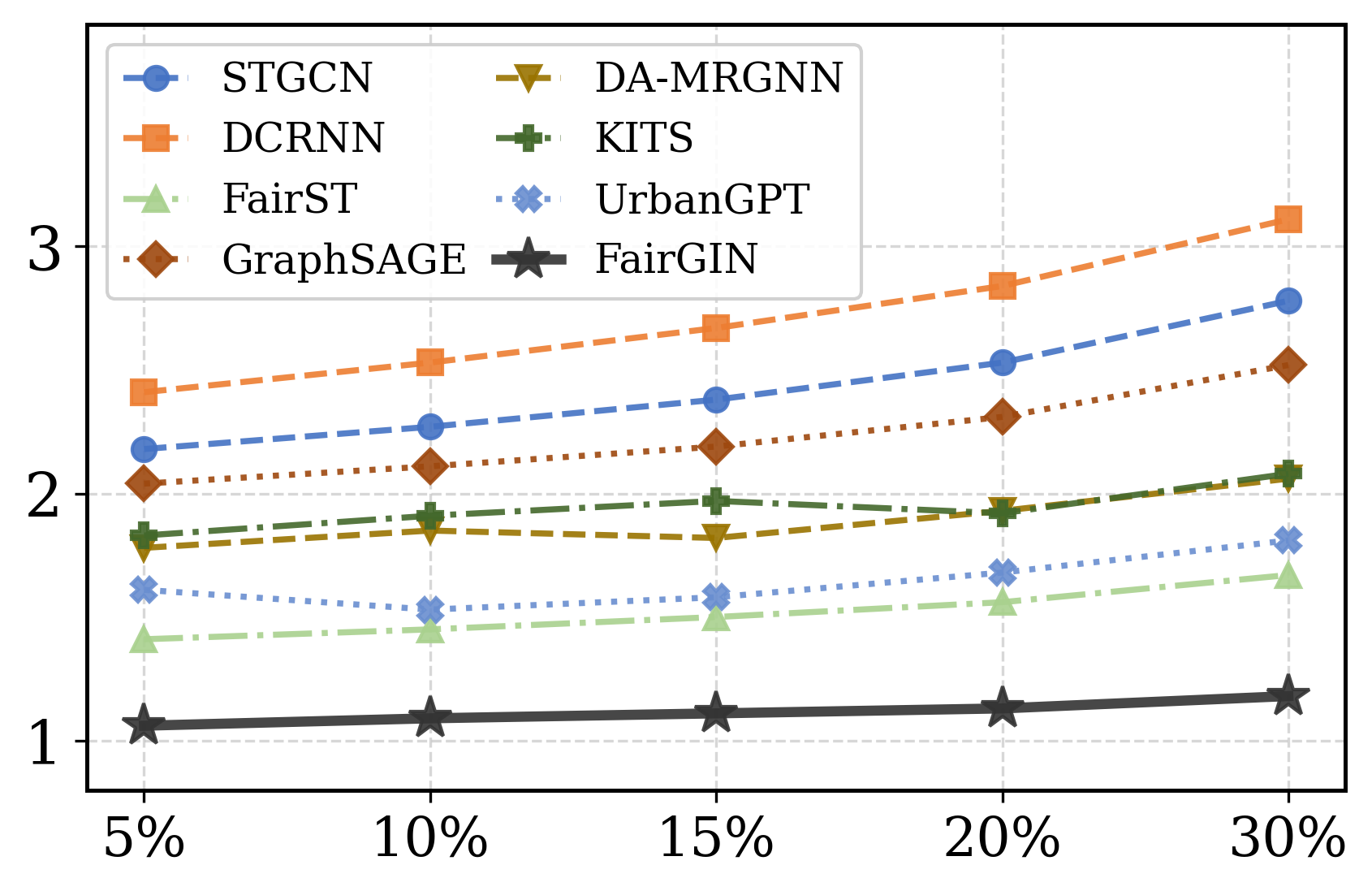}
        \subcaption{}
        \label{fig:subfig_b7}
    \end{minipage}
    \caption{(a) Sensitivity of MAE and RFG to the ESIT mask ratio~$\rho$. (b) Sensitivity of MAE and RFG to the fairness regularization weight~$\lambda_{\mathrm{fair}}$. Spatial distribution of newly added stations under different expansion patterns: (c) localized expansion, (d) regional expansion. Prediction performance of different methods in terms of MAE and RFG under (e) localized expansion and (f) regional expansion. Prediction performance of different methods across expansion rates on NYC Citi Bike, (g) MAE, (h) RFG.}
    \label{fig:my_label4}
\end{figure*}

To examine the contribution of each component in FairGIN, we conduct ablation studies by removing or simplifying key modules from the full framework. As shown in Fig.~\ref{fig:my_label3}, \textbf{w/o ESIT} removes Expansion-Simulated Increment Training, \textbf{w/o KT} replaces attention-based knowledge transfer with standard GCN aggregation, \textbf{w/o FL} sets $\lambda_{\mathrm{fair}} = 0$, and \textbf{w/o TS} replaces the learnable station-specific temperature $\tau_j$ with a fixed value $\tau = 1$. All variants exhibit clear performance degradation on both datasets, confirming that each component contributes to the overall effectiveness of FairGIN. Removing ESIT causes the most substantial decline, with MAE increasing by 27.3\% and RFG increasing by 64.0\% on NYC. This result highlights the importance of explicitly aligning the training process with cold-start deployment conditions, where newly deployed stations lack historical demand observations. The \textbf{w/o KT} variant increases MAE by 17.0\% and RFG by 47.4\%, showing that standard GCN aggregation is insufficient for learning informative representations of new stations and that adaptive knowledge transfer is essential for identifying relevant existing stations. The \textbf{w/o FL} variant has a limited impact on prediction accuracy, with MAE increasing by only 2.2\%, but raises RFG from 1.14 to 1.93. This gap demonstrates that optimizing prediction accuracy alone does not ensure income-stratified fairness across neighborhoods. Finally, the \textbf{w/o TS} variant increases MAE by 8.1\% and RFG by 17.5\%, suggesting that station-specific temperature scaling helps adapt the concentration of attention weights to heterogeneous new-station contexts. Overall, the ablation results show that ESIT provides the foundation for cold-start generalization, while KT, FL, and TS further improve representation quality and fairness-aware prediction.

\subsection{Sensitivity and Robustness Analysis}

\textbf{Parameter sensitivity.}~Figs.~\ref{fig:subfig_a4} and~\ref{fig:subfig_b4} report the sensitivity of FairGIN to the ESIT mask ratio $\rho$ and the fairness regularization weight $\lambda_{\mathrm{fair}}$ on NYC Citi Bike. For $\rho \in \{0.05, 0.10, 0.15, 0.20, 0.25\}$, increasing $\rho$ from 0.05 to 0.15 consistently improves both accuracy and fairness, suggesting that a larger pseudo-new station set introduces more diverse expansion patterns and provides richer supervision across income groups. When $\rho$ exceeds 0.15, performance gradually declines because the observed graph $\mathcal{V}'_A$ becomes increasingly sparse, weakening GCN message passing and reducing the reference pool available for knowledge transfer. This result indicates that an appropriate masking ratio should balance the diversity of simulated expansions against the structural completeness of the observed graph. The best performance is achieved at $\rho = 0.15$, which also falls within the annual expansion rate range of NYC Citi Bike in our dataset, namely 10--20\%. For $\lambda_{\mathrm{fair}} \in \{0.0, 0.1, 0.3, 0.5, 0.8, 1.0\}$ with $\lambda_{\mathrm{sim}} = 0.5$ fixed, increasing $\lambda_{\mathrm{fair}}$ from 0 to 0.3 substantially reduces the fairness gap while introducing only a limited accuracy cost, with MAE increasing by 2.2\%. When $\lambda_{\mathrm{fair}}$ exceeds 0.5, further fairness gains become marginal, whereas the decline in prediction accuracy becomes more evident. This reflects the trade-off between demand prediction accuracy and income-based fairness regularization. We therefore adopt $\rho = 0.15$ and $\lambda_{\mathrm{fair}} = 0.3$ as the default settings.

\textbf{Robustness to expansion patterns.}~Bike-sharing systems may expand through different geographic patterns, ranging from hub-centric localized expansion to city-wide regional expansion. As shown in Figs.~\ref{fig:subfig_c4} and~\ref{fig:subfig_a5}, localized expansion concentrates new stations in a compact area, while regional expansion distributes them proportionally across the five NYC boroughs. To evaluate robustness to these patterns, we construct two NYC Citi Bike test variants. In the regional setting, new stations are sampled proportionally across boroughs. In the localized setting, all $\mathcal{V}_B$ candidates are sampled within a 3\,km radius of a randomly selected centroid and averaged over five seeds. Figs.~\ref{fig:subfig_a7} and~\ref{fig:subfig_a6} compare all baselines under localized and regional expansion, respectively. Under localized expansion, transductive methods, including STGCN and DCRNN, show the largest MAE increases, at 9.4\% and 9.3\%, indicating that fixed-neighborhood assumptions become less reliable when new stations are spatially concentrated. FairST and UrbanGPT remain relatively stable in accuracy, with MAE increases of 4.3\% and 1.0\%, but their fairness gaps increase under the narrower income distribution of localized rollout. DA-MRGNN achieves the strongest baseline accuracy under localized expansion, although its RFG increases by 6.1\%. FairGIN limits the MAE increase to 2.7\% and the RFG increase to 4.4\%, suggesting that temperature-scaled soft attention supports more stable knowledge transfer when the reference pool is geographically constrained.

\begin{table}[t]
\centering
\caption{Deployment equity simulation. G$^-$ ratio denotes the share of selected stations in $\mathcal{G}^-$.}
\label{tab:my_label3}
\setlength{\tabcolsep}{3.0pt}
\small
\begin{tabular}{llcccc}
\toprule
Dataset & Strategy & Top-10 & Top-20 & Top-30 & Mean G$^-$ \\
\midrule
\multirow{4}{*}{NYC}
  & Demand-only & 0.20 & 0.25 & 0.28 & 1.84 \\
  & Random      & 0.47 & 0.46 & 0.48 & 2.31 \\
  & FairGIN     & \textbf{0.60} & \textbf{0.55} & \textbf{0.53} & \textbf{2.47} \\
  & \textit{Base rate} & \multicolumn{3}{c}{\textit{0.52}} & -- \\
\midrule
\multirow{4}{*}{Seattle}
  & Demand-only & 0.18 & 0.22 & 0.27 & 1.61 \\
  & Random      & 0.49 & 0.51 & 0.48 & 2.04 \\
  & FairGIN     & \textbf{0.58} & \textbf{0.53} & \textbf{0.51} & \textbf{2.19} \\
  & \textit{Base rate} & \multicolumn{3}{c}{\textit{0.52}} & -- \\
\bottomrule
\end{tabular}
\end{table}

\textbf{Robustness to different expansion rates.}~Real-world bike-sharing systems expand at different scales. To evaluate the stability of FairGIN under varying expansion sizes, we vary the expansion rate $r \in \{5\%, 10\%, 15\%, 20\%, 30\%\}$ over the candidate new-station set $|\mathcal{V}_B|$ on NYC Citi Bike, with five independent trials for each setting. We compare all baselines except ARIMA and LSTM. Figs.~\ref{fig:subfig_b5} and~\ref{fig:subfig_b7} report MAE and RFG across different expansion rates. This setting allows us to examine whether each method can preserve both predictive accuracy and group-level fairness as the number of unseen stations gradually increases. Graph-based methods, including STGCN and DCRNN, show more noticeable performance degradation as $r$ increases, since a larger set of new stations amplifies the discrepancy between the training and inference graphs. This suggests that models relying on a fixed graph structure are sensitive to expansion-induced topology shifts, especially when many nodes lack historical observations.

As the expansion rate increases, inductive methods generally maintain more stable overall performance than fixed-graph models. GraphSAGE, DA-MRGNN, and KITS show relatively consistent predictive accuracy, although their results still fluctuate across trials due to changes in the sampled new-station composition. UrbanGPT also remains relatively stable in MAE at higher expansion rates, but its fairness gap increases slightly, indicating that prompt-based spatial reasoning alone may not sufficiently correct group-level demand disparities. In comparison, FairGIN consistently achieves the lowest MAE and RFG across all evaluated expansion rates. From $r = 5\%$ to $r = 30\%$, its performance changes by at most 5.6\% in MAE and 11.3\% in RFG. This stability can be attributed to the expansion-simulated training strategy, which exposes the model to diverse pseudo-new stations during training, and the fairness-aware objective, which regularizes group-level prediction errors under varying station compositions.

\subsection{Computational Complexity Analysis}

\begin{table}[t]
\centering
\small
\caption{Complexity comparison ($N\!=\!37$, $M\!=\!16$, $|\mathcal{E}|\!=\!604$; RTX~4090).
  \ding{51}/\ding{55}: cold-start support. Mean\,$\pm$\,std (100 inference / 20 training runs).}
\label{tab:complexity}
\setlength{\tabcolsep}{4pt}
\begin{tabular}{@{}lcccc@{}}
\toprule
\textbf{Model} & \textbf{Cold} & \textbf{\#Params} &
  \textbf{Train (ms)} & \textbf{Infer (ms)} \\
\midrule
ARIMA        & \ding{55}       & $\sim$5/stn & ---             & ---             \\
UrbanGPT     & $\sim$\ding{51} & $\sim$7.2B  & ---             & ---             \\
\midrule
LSTM         & \ding{55}       & 200.7K      & $1.8_{\pm1.0}$  & $0.29_{\pm0.02}$ \\
STGCN        & \ding{55}       & 59.0K       & $3.5_{\pm1.1}$  & $0.95_{\pm0.19}$ \\
DCRNN        & \ding{55}       & 148.7K      & $91.4_{\pm17.2}$ & $51.2_{\pm9.4}$ \\
GraphSAGE    & \ding{51}       & 36.7K       & $1.6_{\pm0.1}$  & $0.41_{\pm0.26}$ \\
DA-MRGNN     & \ding{51}       & 162.2K      & $5.7_{\pm2.7}$  & $1.94_{\pm0.39}$ \\
KITS         & \ding{51}       & 135.6K      & $3.6_{\pm0.6}$  & $1.33_{\pm0.30}$ \\
\textbf{FairGIN} & \ding{51}   & \textbf{211.2K} & $11.3_{\pm4.9}$ & $1.96_{\pm0.63}$ \\
\bottomrule
\end{tabular}
\end{table}

Table~\ref{tab:complexity} compares the parameter count, per-step training time, and per-pass inference latency of FairGIN with all deep-learning baselines on the NYC Citi Bike graph ($N!=!37$, $M!=!16$, $|\mathcal{E}|!=!604$), using an NVIDIA RTX~4090 GPU. FairGIN contains 211.2K trainable parameters, comparable to LSTM (200.7K) and DA-MRGNN (162.2K), with a moderate increase over KITS (135.6K). This additional capacity supports cold-start prediction and income-aware fairness optimization within a unified model. Combining a cold-start model such as KITS with a separate post-hoc fairness module would require additional parameters and inference operations.

FairGIN completes each training step in 11.3,ms and each forward pass in 1.96,ms. Its training time is approximately eight times lower than that of DCRNN (91.4,ms), whose recurrent graph architecture sequentially unrolls $T=24$ graph convolution steps. FairGIN instead encodes historical demand through a GRU before graph propagation, limiting the overhead introduced by the Cayley orthogonal transform and gated knowledge fusion. Its inference latency is comparable to DA-MRGNN (1.94,ms) and approximately 26 times lower than DCRNN (51.22,ms). The Cayley transform requires $\mathcal{O}(d^3)$ operations to construct the orthogonal matrix, but its practical cost remains small at $d=128$, corresponding to approximately 2.1 million floating-point operations. Temperature-scaled attention over the $M\times N$ station matrix adds $\mathcal{O}(MNd)$ complexity, which is also modest for bike-sharing networks of this scale. These results indicate that FairGIN maintains practical computational efficiency while jointly supporting network expansion and fairness-aware prediction. ARIMA does not require GPU computation, but it fits an independent model for each station and has an inference complexity of $\mathcal{O}(Npq)$. UrbanGPT relies on a 7.2B-parameter language-model backbone and requires dedicated server-side infrastructure, making its per-pass latency and deployment cost not directly comparable with the lightweight deep-learning baselines.

\begin{figure}[t]
    \centering
    \begin{minipage}[b]{.24\textwidth}
        \centering
        \includegraphics[width=\linewidth]{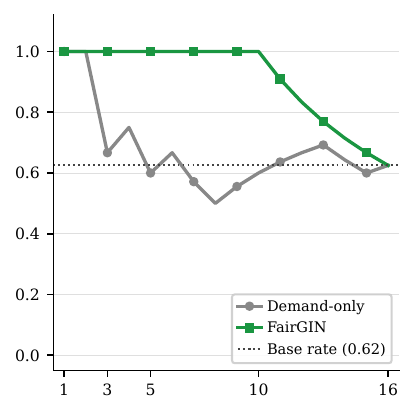}
        \subcaption{}
        \label{fig:subfig_a7}
    \end{minipage}
    \hfill
    \begin{minipage}[b]{.24\textwidth}
        \centering
        \includegraphics[width=\linewidth]{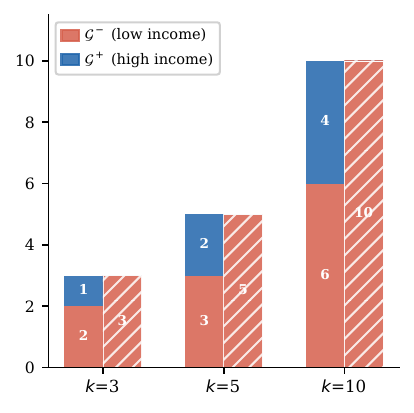}
        \subcaption{}
        \label{fig:subfig_b7}
    \end{minipage}
    \caption{Deployment outcomes on the NYC Citi Bike development split with $M\!=\!16$. Red denotes $\mathcal{G}^{-}$ and blue denotes $\mathcal{G}^{+}$. Panel~(a) reports the proportion of $\mathcal{G}^{-}$ stations in the top-$k$ selections, with the dotted line showing their candidate-set proportion of 0.625. Panel~(b) reports the income-group composition at $k\!=\!3$, $5$, and $10$. Solid and hatched bars represent demand-only and FairGIN selection, respectively.}
    \label{fig:my_label5}
\end{figure}

\subsection{Deployment Equity Simulation}

Table~\ref{tab:my_label3} evaluates three deployment strategies on both datasets. We report the fraction of selected stations belonging to the disadvantaged group $\mathcal{G}^-$, denoted as the G$^-$ ratio, together with the mean predicted demand of the selected $\mathcal{G}^-$ stations under the top-10, top-20, and top-30 selection thresholds. The demand-only strategy selects a substantially smaller share of $\mathcal{G}^-$ stations than their candidate base rate. On NYC, only 0.20 of the top-10 selected stations belong to $\mathcal{G}^-$, compared with a base rate of 0.52, suggesting that ranking stations solely by predicted demand may give insufficient attention to disadvantaged-group locations during deployment planning. This gap also indicates that historical demand patterns alone may not fully reflect the actual service needs of lower-income communities.

Random selection remains close to the candidate base rate, whereas FairGIN consistently increases the G$^-$ ratio across both cities and all selection thresholds, reaching 0.60 for the top-10 stations on NYC and 0.58 on Seattle. Importantly, FairGIN does not improve disadvantaged-group representation by indiscriminately selecting low-demand locations. Compared with random selection, it increases the mean predicted demand of the selected $\mathcal{G}^-$ stations from 2.31 to 2.47 on NYC and from 2.04 to 2.19 on Seattle. This indicates that the model identifies disadvantaged-group locations that remain valuable from a demand perspective, rather than prioritizing fairness at the expense of service efficiency. Overall, these results demonstrate that the deployment scoring strategy improves disadvantaged-group representation while preserving the demand relevance of the selected stations, thereby achieving a more balanced allocation of new bike-sharing infrastructure.

\begin{table}[t]
\centering
\footnotesize
\caption{Representative deployment results on the NYC Citi Bike development split with $\alpha\!=\!0.2$. For $\mathcal{G}^{+}$, $S_j\!=\!\hat{y}_j$. For $\mathcal{G}^{-}$, $S_j\!=\!\hat{y}_j+0.2$. DO and FG denote demand-only and FairGIN selection, respectively. \ding{51}/\ding{55} indicates inclusion or exclusion from the top 10.}
\label{tab:my_label4}
\setlength{\tabcolsep}{3pt}
\begin{tabular}{@{}lcrrcc@{}}
\toprule
\textbf{Station} &
\textbf{Group} &
\multicolumn{1}{c}{$\hat{y}_j$} &
\multicolumn{1}{c}{$S_j$} &
\textbf{DO} &
\textbf{FG} \\
\midrule
\multicolumn{6}{@{}l}{\itshape Selected by both strategies, 2 of 6 shown} \\[1pt]
Sip Ave
& $\mathcal{G}^{-}$ & $+0.028$ & $+0.228$ & \ding{51} & \ding{51} \\
Columbia Park
& $\mathcal{G}^{-}$ & $-0.063$ & $+0.137$ & \ding{51} & \ding{51} \\
\midrule
\multicolumn{6}{@{}l}{\itshape Selected only by demand-only ranking} \\[1pt]
5 Corners Library
& $\mathcal{G}^{+}$ & $+0.007$ & $+0.007$ & \ding{51} & \ding{55} \\
Dixon Mills
& $\mathcal{G}^{+}$ & $-0.010$ & $-0.010$ & \ding{51} & \ding{55} \\
York St
& $\mathcal{G}^{+}$ & $-0.027$ & $-0.027$ & \ding{51} & \ding{55} \\
Astor Place
& $\mathcal{G}^{+}$ & $-0.038$ & $-0.038$ & \ding{51} & \ding{55} \\
\midrule
\multicolumn{6}{@{}l}{\itshape Selected only by FairGIN} \\[1pt]
Lafayette Park
& $\mathcal{G}^{-}$ & $-0.070$ & $+0.130$ & \ding{55} & \ding{51} \\
Communipaw \& Berry
& $\mathcal{G}^{-}$ & $-0.106$ & $+0.094$ & \ding{55} & \ding{51} \\
Dey St
& $\mathcal{G}^{-}$ & $-0.136$ & $+0.064$ & \ding{55} & \ding{51} \\
Danforth Light Rail
& $\mathcal{G}^{-}$ & $-0.183$ & $+0.017$ & \ding{55} & \ding{51} \\
\bottomrule
\end{tabular}
\end{table}

\subsection{Equity-Aware Deployment Case Study}

To complement the aggregate results in Tables~\ref{tab:my_label2} and~\ref{tab:my_label3}, we conduct a station-level case study on the NYC Citi Bike development split. The candidate set contains $M\!=\!16$ new stations, including $|\mathcal{G}^{-}|\!=\!10$ low-income stations and $|\mathcal{G}^{+}|\!=\!6$ high-income stations. We compare demand-only ranking with FairGIN's equity-calibrated deployment score
\begin{equation}
S_j = \hat{y}_j + \alpha \mathbf{1}[j\in\mathcal{G}^{-}],
\end{equation}
where $\alpha\!=\!0.2$ controls the additional weight assigned to stations in $\mathcal{G}^{-}$.

Table~\ref{tab:my_label4} reports the predicted demand $\hat{y}_j$, equity-calibrated score $S_j$, income group, and top-10 selection status of representative candidate stations. For stations in $\mathcal{G}^{+}$, the deployment score remains equal to the original predicted demand. Stations in $\mathcal{G}^{-}$ receive an additional score of $0.2$ under the fairness adjustment. As a result, four $\mathcal{G}^{+}$ stations with scores ranging from $-0.038$ to $0.007$ are replaced by four $\mathcal{G}^{-}$ stations whose adjusted scores range from $0.017$ to $0.130$. Each elevated station therefore ranks above every displaced station under the equity-calibrated deployment criterion. The original demand differences between these stations range from 0.06 to 0.19 normalized units, indicating that the adjustment mainly affects candidates with relatively similar predicted demand rather than promoting stations with consistently low service potential.

Figs.~\ref{fig:my_label5}(a) and (b) examine how the two strategies behave under different deployment budgets. Under demand-only ranking, the proportion of selected $\mathcal{G}^{-}$ stations falls to 0.60 at both $k\!=\!5$ and $k\!=\!10$, slightly below their candidate-set proportion of 0.625. With $\alpha\!=\!0.2$, FairGIN selects only $\mathcal{G}^{-}$ stations for all $k\!\leq\!10$. This result demonstrates that the equity-calibrated score can substantially increase the representation of low-income areas. It also shows that the choice of $\alpha$ directly determines the balance between income-group representation and demand-based ranking. In practice, this parameter should therefore be selected according to the intended equity target and deployment constraints.
\section{Discussion and Limitations}

This section discusses and interprets the main findings through four research questions that provide broader insights beyond the quantitative analysis presented in Section~VI.

\textbf{RQ1: Does integrating cold-start prediction with fairness regularization improve equity beyond accuracy-focused inductive methods?}

The results in Table~\ref{tab:my_label2} show that inductive capability alone does not eliminate income-based prediction disparities. On the NYC Citi Bike dataset, KITS achieves the lowest MAE among the baselines, but its RFG remains 2.19, only slightly lower than the 2.31 obtained by the transductive DCRNN model. This finding suggests that generalization to unseen stations does not necessarily produce equitable predictions because transferred knowledge may preserve income-related patterns embedded in historical ridership data.

FairGIN addresses this issue during representation learning rather than through an output-level correction. Income-stratified supervision is incorporated into the same forward pass used to construct cold-start representations, allowing the fairness objective to influence new-station embeddings before demand prediction. The ablation results support this design. Removing $\mathcal{L}_{\mathrm{fair}}$ increases RFG by 0.47 on NYC, while MAE increases by only 0.08. Fairness can therefore be improved without a substantial reduction in predictive accuracy when it is integrated directly into cold-start representation learning. By comparison, a separate fairness correction applied after prediction cannot modify the income-related bias already encoded in the spatial representation.

\textbf{RQ2: Why does feature heterogeneity at inference time disadvantage low-income new stations?}

At inference time, new stations are represented only by spatial attributes, including POI density, transit connectivity, and surrounding land use, whereas existing stations also provide complete historical demand sequences. Knowledge transfer therefore depends on the similarity between the spatial representation of a new station and the learned representations of existing stations. In New York City, high-demand stations are often located in higher-income areas with dense amenities and stronger transit access. These characteristics are also reflected in the spatial features used by the model. New stations in lower-income and less amenity-dense areas may consequently receive weaker attention from the most informative existing stations, producing transferred representations that underestimate their latent demand.

FairGIN reduces this bias through orthogonal alignment and gated fusion. The Cayley transform adjusts the transferred representation to reduce its dependence on income-related spatial patterns. The gated fusion mechanism then allows each new station to balance the aligned transferred information with its own spatial embedding. The sensitivity analysis in Fig.~\ref{fig:my_label4}(b) supports this interpretation. Increasing $\lambda_{\mathrm{fair}}$ from 0.0 to 0.3 reduces RFG by 33.3\%, while MAE increases by only 2.2\%. This result indicates that much of the fairness gap arises from correctable representation bias rather than an unavoidable limitation of cold-start prediction.

\begin{table}[t]
\centering
\normalsize
\caption{MAE variation within the income groups for the $M\!=\!16$ new stations in the NYC development split. The binary grouping masks differences in prediction error within $\mathcal{G}^{-}$. $^\dagger$ indicates that no ACS income record is available for the corresponding census tract.}
\label{tab:my_label5}
\setlength{\tabcolsep}{6pt}
\renewcommand{\arraystretch}{1.08}
\begin{tabular}{@{}lccc@{}}
\toprule
\textbf{Subgroup} & \textbf{$n$} & \textbf{Income range} & \textbf{MAE} \\
\midrule
$\mathcal{G}^{+}$, high income
& 6 & \$90k to \$149k & 0.343 \\
\midrule
$\mathcal{G}^{-}$, near median
& 3 & \$60k to \$85k & 0.588 \\
$\mathcal{G}^{-}$, low income
& 2 & \$36k to \$59k & 0.717 \\
$\mathcal{G}^{-}$, data sparse$^\dagger$
& 5 & Not available & 1.586 \\
\cmidrule(l){2-4}
$\mathcal{G}^{-}$, overall
& 10 & Not applicable & 1.113 \\
\bottomrule
\end{tabular}
\end{table}

\textbf{RQ3: How well does FairGIN generalize across urban contexts and expansion scales?}

The Seattle experiments show that FairGIN remains effective in a city with a different demographic and spatial structure. Seattle has a higher median household income than New York City, with values of \$93,500 and \$72,800, respectively, and exhibits different patterns of income segregation. Despite these differences, FairGIN reduces RFG by 31.2\% relative to the strongest Seattle baseline, close to the 33.3\% reduction achieved on NYC. This consistency suggests that the fairness objective is not tied to a specific income distribution and can address group-level disparities across different urban settings.

The expansion-rate experiments in Fig.~\ref{fig:my_label4}(g--h) further demonstrate the robustness of FairGIN. At $r=30\%$, its MAE variation across random seeds remains within 5.6\%, while its stability advantage over the baselines becomes more pronounced. This robustness stems from ESIT, which exposes the model to pseudo-new station sets of varying sizes and compositions during training, reducing sensitivity to specific expansion patterns. At $r=5\%$, FairGIN maintains a clear RFG advantage over KITS, showing that fairness supervision remains effective with fewer newly added stations.

\textbf{RQ4: What are the main limitations and directions for future work?}

\textit{Binary income partitioning.}~FairGIN divides stations into two income groups, $\mathcal{G}^{+}$ and $\mathcal{G}^{-}$, using the city-wide median income as the threshold. Although this binary formulation provides a clear group-fairness objective, it does not capture differences in income disadvantage within $\mathcal{G}^{-}$.

Table~\ref{tab:my_label5} examines this within-group variation for the $M\!=\!16$ new stations in the NYC development split. Among stations with available income records, MAE increases from 0.588 for the near-median subgroup (\$60k to \$85k) to 0.717 for the low-income subgroup (\$36k to \$59k). The five stations without ACS income records have a substantially higher MAE of 1.586, raising the overall $\mathcal{G}^{-}$ MAE to 1.113. These results show that the binary grouping masks meaningful variation in prediction difficulty. Future work could use income quintiles or a continuous sensitive attribute, such as Wasserstein fairness~\cite{jiang2020wasserstein}, and assign greater fairness weights to stations facing greater income disadvantage.

\textit{Further limitations.}~The current study is subject to several additional limitations. Neighborhood income is treated as the only protected attribute, although race, car ownership, and proximity to public transport may exert effects that are not fully explained by income alone. A multi-attribute fairness formulation could therefore provide a more comprehensive account of overlapping sources of disadvantage. The spatial encoder also relies on POI and demographic features captured at a single point in time. As neighborhood conditions evolve after station deployment, periodically updating these inputs may help preserve long-term accuracy and equity without requiring full model retraining. In addition, both datasets are drawn from docked bike-sharing systems in the United States and benefit from relatively dense ACS coverage. Whether FairGIN generalizes to dockless networks, cities where income is less strongly associated with mobility patterns, or data-scarce regions in the Global South remains to be investigated.
\section{Related Work}

This section reviews six key strands of literature relevant to our work. We first provide an overview of urban mobility systems and bike-sharing in sustainable low-carbon transport. We then examine bike-sharing expansion, highlighting existing limitations in modeling newly added stations. Next, we review fairness-aware urban mobility studies, focusing on equity in demand prediction, followed by a discussion of fairness in machine learning and graph neural networks that situates FairGIN's fairness objectives within the broader data mining literature. We then discuss spatiotemporal graph neural networks for mobility prediction, emphasizing methods for dynamic and evolving network structures. Finally, we review knowledge transfer and cold-start prediction techniques, situating the attention-based knowledge transfer mechanism of FairGIN within this broader context.

\subsection{Urban Mobility Systems}

As cities seek to reduce transportation-related carbon emissions, shared mobility systems, particularly bike-sharing, have become an important component of sustainable urban transport infrastructure~\cite{amatuni2020does}. By offering flexible, zero-emission options for short-distance travel and last-mile connectivity, bike-sharing systems can reduce reliance on private vehicles and support urban decarbonization goals~\cite{liang2023cross}. Realizing these benefits, however, requires addressing key operational challenges, including station rebalancing and optimization \cite{liu2015station}, spatio-temporal demand prediction~\cite{zhao2016predicting}, and user behavior modeling~\cite{chen2020predicting}. Sathishkumar et al. \cite{sathishkumar2020using} analyzed multi-year real-world data to identify spatio-temporal usage patterns and evaluate system performance across diverse urban contexts. Most existing studies focus on operational efficiency at established stations, typically assuming fixed network topology and stable historical usage patterns. Such assumptions limit their applicability to expanding bike-sharing networks, where new stations continuously reshape spatial structure and mobility demand. Recent studies have examined expansion-related issues, including optimal station placement based on spatial coverage, urban morphology, and built environment characteristics~\cite{duran2021demand}, as well as the effects of system expansion on transit integration and heterogeneous demand patterns~\cite{zhu2022approaching, bi2021analysis, chen2021dynamic}. Despite these advances, predicting how newly deployed stations affect demand dynamics across the existing network remains insufficiently explored, especially for equitable low-carbon mobility expansion. More recently, data-driven urban computing has shifted toward large-scale foundation model approaches that leverage broad pretraining to support generalized inference across diverse urban tasks~\cite{li2024urbangpt}. While such methods expand the applicability of urban prediction models, they typically operate on established network configurations and do not explicitly address the distributional shift that arises when new stations with no demand history are integrated into an expanding graph.

\subsection{Bike-Sharing System Expansion}

Significant research efforts have targeted the prediction of optimal expansion strategies for bike-sharing systems, including the deployment of new stations~\cite{8640047,liang2024time} and capacity adjustments at existing facilities~\cite{liang2023cross,liang2023deep}. Common prediction methods involve spatial clustering, demand mapping, and various optimization techniques, often guided by historical usage data or static spatial heuristics. However, many existing studies assume that newly added stations exhibit demand characteristics similar to those of existing locations, neglecting the explicit prediction of temporal shifts and relational dynamics resulting from system expansion. Such assumptions significantly reduce model adaptability in real-world scenarios, where new stations may serve entirely different urban functions, land-use types, or demographic groups. Liu et al.~\cite{liu2017functional} introduced a hierarchical zone-based demand prediction model designed to estimate average demand at newly deployed stations across different stages of expansion. Although our approach leverages contextual information similarly, it differs fundamentally by explicitly modeling dynamic network evolution and the changing spatial-temporal interactions among stations. Furthermore, our method simultaneously predicts both instantaneous and expected demands at a fine-grained temporal resolution, rather than merely aggregating demand at a coarse zone level or assuming station-level homogeneity. More recent inductive approaches, including KITS~\cite{xu2025kits} and DA-MRGNN~\cite{liang2023cross}, have extended graph-based prediction to newly added stations by simulating cold-start conditions or transferring demand patterns across graph domains. These methods improve generalization to unseen stations and represent important steps toward operational deployment in expanding networks. However, they focus exclusively on predictive accuracy and do not account for the income-based disparities that may be embedded in or amplified by such predictions. Without fairness-aware supervision, demand estimates for new stations risk systematically reflecting and perpetuating existing socioeconomic inequities in infrastructure access.

\subsection{Fairness in Urban Mobility}

Equity in urban transportation has received increasing attention, as empirical studies show that shared mobility services often reproduce existing socioeconomic disparities rather than alleviating them. In docked bike-sharing systems, higher-income and more educated residents tend to have greater spatial access to stations \cite{hosford2018public, giuffrida2023social}, while underserved neighborhoods remain systematically under-provisioned \cite{mooney2019freedom, mohiuddin2023does}. These patterns reflect the interaction between deployment decisions, unequal land use, infrastructure investment, and service coverage, creating risks that data-driven models may further amplify. Recent studies have therefore introduced fairness-aware methods for mobility prediction. Zheng et al.~\cite{zheng2023fairness} propose SA-Net, which integrates sociodemographic features and bias-mitigation regularization for ride-hailing demand forecasting. Xia et al.~\cite{xia2025fairtp} develop FairTP to reduce long-term regional disparities in traffic prediction, while Zhuang et al.~\cite{zhuang2025mitigating} design a residual-aware spatiotemporal GNN with an equality-enhancing loss for urban demand forecasting. More broadly, the machine learning fairness literature has established a taxonomy of fairness criteria that includes group fairness, which requires equitable aggregate outcomes across demographic groups, and individual fairness, which requires similar predictions for comparably situated instances~\cite{mehrabi2021survey}. These distinctions motivate the complementary use of the Region-based Fairness Gap and the Individual-based Fairness Gap in our evaluation, capturing both population-level and station-level prediction disparities. Despite their contributions, existing fairness-aware mobility methods share a common limitation: they are designed for fixed network settings in which all stations are observed during training. None addresses the cold-start scenario where new stations must be evaluated at inference time without any prior demand history, leaving the problem of equitable demand prediction for newly deployed stations unresolved.

\subsection{Fairness in Machine Learning}

Fairness-aware machine learning has established two dominant paradigms for constraining model behavior with respect to sensitive attributes. Group fairness requires equitable aggregate outcomes across demographic groups, with demographic parity and equalized odds as prominent operationalizations~\cite{hardt2016equality, mehrabi2021survey}. Individual fairness~\cite{dwork2012fairness} requires that similarly situated instances receive similar predictions, a principle that motivates station-level as well as population-level disparity evaluation; these two paradigms directly underpin the complementary Region-based Fairness Gap and Individual-based Fairness Gap metrics used in our work. In graph neural networks, enforcing fairness introduces additional challenges because neighborhood aggregation can encode and propagate sensitive attribute correlations across the graph structure~\cite{dai2021say, ma2021subgroup}. Methods such as FairGNN~\cite{dai2021say} address this through adversarial debiasing, while subgroup-constrained objectives~\cite{ma2021subgroup} target generalization disparity across demographic subgroups. However, all of these approaches assume a static transductive setting in which node labels, group annotations, and neighborhood structure are fully observable at training time, and none is designed for expanding graphs where newly deployed nodes are absent during training and must be evaluated under cold-start conditions.

\subsection{Dynamic Graph Modeling}

Dynamic graph prediction has been widely studied for modeling spatial and temporal dependencies in transportation systems. Early approaches combine recurrent neural networks with graph convolutions for sequential prediction~\cite{10.5555/3304222.3304273,roy2021unified}, but often suffer from high computational costs and gradient instability over long sequences. CNN-based methods replace recurrent structures with one-dimensional convolutions to improve efficiency~\cite{10.5555/3304222.3304273}, although repeated graph propagation may oversmooth fine-grained station relationships. More recent studies employ adaptive graph structures~\cite{zheng2023spatio} and transformer-based architectures~\cite{jin2023spatio} to capture changing connectivity and long-range dependencies. Inductive models such as GraphSAGE~\cite{hamilton2017inductive} generalize to unseen nodes through neighborhood aggregation, while IGNNK~\cite{DBLP:journals/corr/abs-2006-07527} reconstructs spatiotemporal signals from randomly sampled subgraphs. Dynamic graph representation learning further models evolving node states and edge formations~\cite{JMLR:v21:19-447}, and architectures such as PDFormer~\cite{jiang2023pdformer} improve traffic prediction through adaptive spatial and temporal attention. However, most existing methods assume a fixed and fully observed node set, leaving unresolved the feature heterogeneity that arises when newly deployed stations lack historical demand observations. Our framework addresses this limitation by updating graph connectivity at each time step according to spatial proximity, functional similarity, and temporal demand variation, thereby improving prediction under bike-sharing network structures.

\subsection{Knowledge Transfer for Cold-Start Prediction}

Knowledge transfer addresses data scarcity at cold-start nodes by leveraging representations from data-rich sources to support inference at data-sparse targets, a challenge formalized across instance, feature, and model transfer paradigms by Pan and Yang~\cite{pan2009survey}. In the graph domain, GraphSAGE~\cite{hamilton2017inductive} introduced inductive representation learning through neighborhood sampling and aggregation, enabling embedding generation for nodes absent during training. Cold Brew~\cite{zheng2021cold} extended this by distilling graph-aware representations into a feature-only student model, explicitly targeting the strict cold-start scenario in which a node has no available neighbors at inference time. Spatiotemporal prediction has seen parallel efforts: IGNNK~\cite{DBLP:journals/corr/abs-2006-07527} reconstructs demand signals at unobserved locations via random subgraph kriging over source nodes; STEP~\cite{shao2022pre} enhances spatiotemporal GNNs through long-term temporal pre-training to improve generalization under limited observations; and ST-GFSL~\cite{lu2022spatio} addresses few-shot spatiotemporal prediction via cross-city meta-learning and node-level parameter matching. Most directly relevant to the expanding network setting, KITS~\cite{xu2025kits} combines inductive kriging with an increment training strategy that simulates cold-start conditions before deployment. Despite these advances, existing methods transfer knowledge through uniform neighborhood aggregation or fixed structural priors, without adapting the weighting of the transfer to the pairwise representational similarity between individual source and target nodes. FairGIN addresses this gap through temperature-scaled soft attention and orthogonal embedding alignment, enabling fine-grained, similarity-aware knowledge transfer under the feature heterogeneity that characterizes expanding bike-sharing networks.
\section{Conclusion}

We proposed FairGIN, a fairness-aware dynamic graph neural network framework for demand prediction in expanding bike-sharing systems, aiming to support more equitable and low-carbon urban mobility. FairGIN jointly addresses two structurally coupled challenges, the inductive prediction gap for newly deployed stations and the income-based bias inherited from historical demand data. Its three core components, namely Expansion-Simulated Increment Training, Attention-Based Knowledge Transfer, and Fairness-Aware Training with deployment scoring, target distinct limitations of prior approaches while forming a unified framework that can be trained in an end-to-end manner. Experiments on NYC Citi Bike and Seattle Bikeshare show that FairGIN achieves leading prediction accuracy while substantially reducing demand disparities across income groups, demonstrating that accurate and fair prediction for newly deployed stations can be achieved simultaneously in expanding bike-sharing networks.

Beyond its technical contributions, this work highlights the importance of fairness-aware modeling in low-carbon urban planning. As data-driven tools increasingly inform transportation investment, models that uncritically reproduce historical inequities may further concentrate the benefits of sustainable mobility in already well-served areas. FairGIN demonstrates that deliberate fairness design at the embedding level can translate model-level equity into more inclusive green mobility access for underserved communities, and we hope it serves as a foundation for broader equity-aware spatiotemporal learning in multi-modal low-carbon networks.

\bibliographystyle{IEEEtran}
\bibliography{references}

\end{document}